\documentclass{article} % For LaTeX2e
\usepackage[preprint]{iclr2027_conference}
\usepackage{times}

\usepackage{amsmath,amsfonts,bm}

\def\eqref#1{equation~\ref{#1}}
\def\1{\bm{1}}

\DeclareMathAlphabet{\mathsfit}{\encodingdefault}{\sfdefault}{m}{sl}
\SetMathAlphabet{\mathsfit}{bold}{\encodingdefault}{\sfdefault}{bx}{n}

\usepackage{hyperref}
\usepackage{url}

\usepackage{graphicx}
\usepackage{amsmath}
\usepackage{booktabs}
\usepackage{lipsum}
\usepackage{multirow}
\usepackage{subcaption}
\usepackage{makecell}
\usepackage{wrapfig}
\usepackage[capitalize,noabbrev]{cleveref}
\usepackage{enumitem}
\usepackage{pifont}

\usepackage{etoc}

\usepackage[nottoc]{tocbibind}
\settocbibname{References}
\usepackage{minitoc}
\usepackage{bbm}

\newcommand{\cmark}{\ding{51}}  % ✓
\newcommand{\xmark}{\ding{55}}  % ✗

\title{$\alpha$Transfer: Coefficient Transfer for Efficient Model Merging}

\renewcommand{\footnotemark}{}

\author{
Shih-Cheng Huang$^{1*}$\thanks{$^*$Equal contribution.}\thanks{Corresponding authors: Shih-Cheng Huang \texttt{<andy.huang@appier.com>}, Zhi Rui Tam \texttt{<ray.eed08g@nctu.edu.tw>}.}
\quad
Zhi Rui Tam$^{1,2*}$
\quad
Chieh-Yen Lin$^{1}$
\quad
Yun-Nung Chen$^{2}$
\\[0.6ex]
\hspace{0.1pt}
\textbf{Hung-yi Lee}$^{2}$
\quad
\textbf{Shao-Hua Sun}$^{1,2}$
\\[1.0ex]
$^1$Appier AI Research
\quad
$^2$National Taiwan University
}

\begin{document}

\doparttoc % Tell to minitoc to generate a toc for the parts
\faketableofcontents % Run a fake tableofcontents command for the partocs

\maketitle

\begin{abstract}
    Model merging offers a promising solution for combining multiple fine-tuned checkpoints into a single model through parameter arithmetic. However, finding optimal merging coefficients requires an extensive search that becomes prohibitively expensive as models scale in both size and number, due to high memory requirements and combinatorial growth in the search space. We show that, within the same model family, models exhibit highly congruent performance distributions over merging coefficients across different model sizes. This distributional similarity enables a practical paradigm we call \textit{$\alpha$Transfer}: searching for optimal coefficients on a small proxy model, then directly transfer them to larger target models. We verify $\alpha$Transfer across multiple merging methods, model families, and tasks. Experimental results demonstrate a 6$\times$ speedup and 70\% memory reduction on vision transformers, and a 20$\times$ speedup and 85\% memory reduction on large language models, while maintaining comparable performance. Our findings establish $\alpha$Transfer as an efficient and generalizable approach to scaling model merging.
\end{abstract}

\section{Introduction}

\begin{wrapfigure}[19]{r}{0.55\linewidth}
    \centering
    \includegraphics[width=\linewidth]{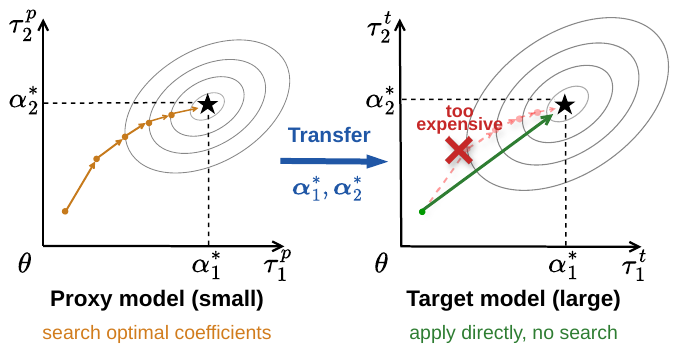}
    \caption{\textbf{Overview of $\alpha$Transfer.} Axes represent task vectors ($\tau_1, \tau_2$); superscripts $p$, $t$ denote proxy and target models. Contours show the performance landscape when merging with coefficients ($\alpha_1, \alpha_2$). Optimal coefficients found on a cheap proxy model transfer directly to the large target, skipping the expensive search.}
    \label{fig:example}
\end{wrapfigure}

The pre-train and fine-tune paradigm, where a powerful pre-trained model is fine-tuned on task-specific data to obtain specialized expert models, has become the dominant approach in deep learning \citep{howard2018universal, radford2019language, bommasani2021opportunities, zhuang2020comprehensive}. This has led to a growing ecosystem of publicly available fine-tuned checkpoints \citep{wolf2020transformers}.
However, real-world applications increasingly demand models capable of handling multiple tasks or integrating diverse skills simultaneously. Maintaining separate fine-tuned models for each capability thus becomes impractical: it incurs significant storage and deployment overhead \citep{fifty2021efficiently, zhang2021survey}, and the knowledge and skills of these isolated models cannot be easily composed to serve diverse applications \citep{ruder2017overview, ni2023does}.

Model merging has emerged as a promising solution to these limitations, directly combining multiple fine-tuned models in weight space into a single unified model capable of handling multiple tasks. Early work demonstrated the feasibility of this approach by merging task-specific models through weighted averaging \citep{jin2023dataless}. Merging was further reframed through the lens of task vectors, the parameter differences between fine-tuned and pre-trained models that capture task-specific knowledge \citep{ilharco2023editing}. In this formulation, the merged model is obtained by adding a scaled combination of task vectors to the pre-trained model, where scaling coefficients determine each task's contribution. Subsequent works proposed various ways to process task vectors during merging, such as resolving sign conflicts \citep{yadav2023ties} or applying sparsity constraints \citep{yu2024language, davari2024model}. Given the significant impact of scaling coefficients on merging performance, other works focus specifically on finding better coefficients, either by learning them adaptively \citep{yang2024adamerging, touayouch2026divmerge} or searching over larger spaces \citep{akiba2025evolutionary}. This line of work has further extended to LLMs, demonstrating skill composition across diverse capabilities such as mathematical reasoning, code generation, and multilingual ability \citep{yu2024language, akiba2025evolutionary, yu2025extend, huang2024chat}.

Despite its promise, \textbf{searching for the merging coefficients} remains a central challenge that limits the scalability of model merging~\citep{he2026mergebench}. Since each candidate set of coefficients requires a full model evaluation to assess, the search cost grows exponentially with the number of tasks, making exhaustive search infeasible. Existing methods address this through two broad strategies: evaluation-based search and gradient-based optimization. The first relies on approximate strategies such as assuming equal coefficients \citep{ilharco2023editing, yadav2023ties, yu2024language}, heuristic search \citep{akiba2025evolutionary}, or Bayesian optimization \citep{lee2025dynamic, liu2025maximizing} to reduce the number of evaluations. However, each evaluation becomes increasingly costly as model size grows, and since these methods still require many evaluations to find good coefficients, the total cost can be substantial for large models. The second strategy avoids repeated evaluation by optimizing coefficients directly against a proxy objective \citep{yang2024adamerging, touayouch2026divmerge}, but requires computing gradients over multiple models during optimization. The memory demand thus scales with both the number of tasks and model size, rendering these methods infeasible when GPU memory is insufficient. Together, these two limitations pose a fundamental barrier to applying model merging in practical, large-scale settings.

The high computation cost of evaluation-based search and the prohibitive memory demand of gradient-based optimization together make coefficient search the primary bottleneck for applying model merging at large scale. A natural question then arises: can we perform this expensive search on a much smaller model instead? We propose \textit{$\alpha$Transfer} (see \cref{fig:example}), which searches for optimal merging coefficients on a small proxy model from the same family and directly applies them to the large target model. Our intuition is that optimal merging coefficients encode both the correlations among tasks and their relative importance, properties largely determined by the tasks themselves and the model family rather than by model scale, and are thus transferable across scales within the same family. We substantially reduce both computation and memory cost by performing coefficient search on a significantly smaller proxy model, making model merging practical even at large scale.

We validate $\alpha$Transfer through extensive experiments on vision and language models, covering six merging methods (Task Arithmetic \citep{ilharco2023editing}, TIES \citep{yadav2023ties}, DARE \citep{yu2024language}, AdaMerging \citep{yang2024adamerging}, AdaMerging++ \citep{yang2024adamerging}, and DivMerge \citep{touayouch2026divmerge}) across CLIP-ViT \citep{radford2021learning}, SigLIP \citep{zhai2023sigmoid} , and Qwen3 \citep{yang2025qwen3} model families. Empirical analysis shows that models within the same family exhibit highly correlated performance landscapes across different coefficient combinations, supporting the feasibility of coefficient transfer. In many cases, transferred coefficients match direct search on the target model with negligible accuracy loss, while achieving up to \textbf{6$\times$} speedup for search-based methods and \textbf{70\%} memory reduction for gradient-based methods on vision transformers, directly addressing the two bottlenecks identified above, and up to \textbf{20$\times$} speedup and \textbf{85\%} memory reduction on large language models.

\section{Related work}
\subsection{Coefficient search in model merging}
While model merging encompasses a broad range of techniques~\citep{yang2026model}, we focus on the core challenge of determining scaling coefficients. We systematically categorize coefficient search into two paradigms: evaluation-based and gradient-based methods, each constrained by distinct bottlenecks in computation and memory, respectively.

\paragraph{Evaluation-based methods.} These methods search the coefficient space by repeatedly evaluating candidate merged models. The simplest variants use uniform coefficients across tasks~\citep{ilharco2023editing,yadav2023ties,yu2024language}, which is cheap but often suboptimal, as per-task or per-layer adjustment substantially improves performance~\citep{yang2024adamerging}. More sophisticated approaches reduce the number of evaluations through evolutionary search~\citep{akiba2025evolutionary} or Bayesian optimization~\citep{lee2025dynamic,liu2025maximizing}. Despite these advances, the cost per evaluation grows with model size, making the total compute budget substantial at scale.

\paragraph{Gradient-based methods.} These methods instead optimize coefficients directly against a proxy objective. AdaMerging~\citep{yang2024adamerging} and AdaMerging++ minimize prediction entropy on unlabeled test data, while DivMerge~\citep{touayouch2026divmerge} uses Jensen-Shannon divergence between merged and individual fine-tuned models. These methods avoid repeated evaluation, but require holding multiple fine-tuned models in memory simultaneously, with memory cost scaling jointly with task count and model size.

Across both categories, all existing approaches conduct the search on the target model itself. We ask whether the search can instead be conducted on a smaller proxy model, enabling existing methods to be applied at scales where they would otherwise be impractical.

\subsection{Searching on small proxies for large targets}
Searching on a small proxy and applying the result to a large target has appeared in several settings outside model merging. $\mu$Transfer~\citep{yang2021tuning} transfers training hyperparameters such as learning rate from a small model to a large one via the $\mu$P parameterization. DoReMi~\citep{xie2023doremi} uses a 280M proxy to determine pretraining data mixture weights for training an 8B model. \citet{khaddaj2025small} characterize how training data influences model behavior consistently across scale, enabling proxy-based data attribution and dataset selection. These works share a common philosophy with ours: quantities optimized on a small model can remain meaningful when transferred to a larger one. However, to our knowledge, this paradigm has not been investigated in model merging. We are the first to show that merging coefficients exhibit such cross-scale transferability within model families, across architectures and merging methods.

\section{$\alpha$Transfer} \label{sec:coeff_trans}
\subsection{Problem formulation}
Consider a model merging problem where we are given a pre-trained model and $T$ task-specific fine-tuned models derived from it. The goal is to combine these models into a single unified model that performs well across all $T$ tasks, without access to the original training data.

Formally, let $\theta_0 \in \mathbb{R}^d$ denote the parameters of the pre-trained model, and let $\theta_1, \ldots, \theta_T$ denote the parameters of the $T$ fine-tuned models. The \textit{task vector} for task $i$ is defined as the parameter difference between the fine-tuned and pre-trained models:
\begin{equation}
    \tau_i = \theta_i - \theta_0.
\end{equation}

\paragraph{Unified formulation.} While existing model merging methods appear diverse in their mechanisms, we observe that they share a common underlying structure: they edit task vectors through various strategies (trimming, masking, etc.), then combine them with learnable or searched coefficients. This insight motivates us to propose a unified formulation that captures this shared structure, enabling systematic analysis and comparison across methods (see~\cref{tab:methods}).
We express a general family of model merging methods as:
\begin{equation}
    \theta_{\text{merged}} = \mathcal{M}(\theta_0, \{\tau_i\}_{i=1}^T; \boldsymbol{\alpha}),
    \label{eq:general}
\end{equation}
where $\mathcal{M}$ is a merging operator that encapsulates both the editing strategy and the combination mechanism, and $\boldsymbol{\alpha}$ denotes the merging coefficients. The key intuition is that most merging methods operate in two conceptual stages: an editing stage where a function $f(\cdot)$ (embedded within $\mathcal{M}$) transforms the task vectors into edited representations $\tilde{\tau}$, followed by a weighting stage where $\boldsymbol{\alpha}$ scales the edited representations and adds them back to $\theta_0$. The granularity at which $\boldsymbol{\alpha}$ is defined gives rise to three common settings:
\begin{itemize}
    \item \textbf{Global:} The merging operator produces a single edited task vector $\tilde{\tau} = f(\{\tau_i\}_{i=1}^T)$, where $f: \mathbb{R}^{T \times d} \rightarrow \mathbb{R}^d$. A single scalar $\alpha \in \mathbb{R}$ is shared across all tasks:
    \begin{equation}
        \theta_{\text{merged}} = \theta_0 + \alpha \cdot \tilde{\tau}.
    \end{equation}
    \item \textbf{Task-wise:} The merging operator produces task-specific edited representations $\tilde{\tau}_i = f_i(\{\tau_j\}_{j=1}^T)$, where $f_i: \mathbb{R}^{T \times d} \rightarrow \mathbb{R}^d$ may leverage all task vectors to compute the representation for task $i$. A distinct scalar $\alpha_i \in \mathbb{R}$ is assigned to each task:
    \begin{equation}
        \theta_{\text{merged}} = \theta_0 + \sum_{i=1}^T \alpha_i \cdot \tilde{\tau}_i.
    \end{equation}
    \item \textbf{Layer-wise:} The merging operator produces task-specific and layer-specific edited representations $\tilde{\tau}_{i,l} = f_{i,l}(\{\tau_{j,l}\}_{j=1}^T)$, where $f_{i,l}: \mathbb{R}^{T \times d_l} \rightarrow \mathbb{R}^{d_l}$ may leverage all task vectors at layer $l$ to compute the representation for task $i$. A distinct scalar $\alpha_{i,l} \in \mathbb{R}$ is assigned to each layer of each task, and the parameters at each layer $l$ are merged independently:
    \begin{equation}
        \theta_{\text{merged}}^{(l)} = \theta_0^{(l)} + \sum_{i=1}^T \alpha_{i,l} \cdot \tilde{\tau}_{i,l}, \quad \forall l = 1, \ldots, L.
    \end{equation}
\end{itemize}

Notably, some recent methods, such as WUDI-Merging~\citep{cheng2025whoever}, are designed to be coefficient-free. They directly obtain the merged weights via optimization without any scaling coefficient. Such methods fall outside our formulation, as no coefficient search is involved, and are therefore not applicable to $\alpha$Transfer. We mark the compatibility of each method in the last column of~\cref{tab:methods}.

\begin{table}[t]
    \centering
    \caption{\textbf{Representative model merging methods under our unified formulation.} Methods are characterized along three dimensions: editing function $f$, coefficient search strategy, and coefficient granularity. The last column indicates compatibility with $\alpha$Transfer: \cmark{} denotes methods whose coefficients are selected via search and are thus compatible, and \xmark{} denotes coefficient-free methods, where no search is involved.}
    \label{tab:methods}
    \setlength{\tabcolsep}{4pt}
    \resizebox{\linewidth}{!}{\begin{tabular}{llllc}
        \toprule
        \textbf{Method} & \textbf{Editing Function} $\boldsymbol{f}$ & \textbf{Coeff. Search} & \textbf{Granularity} & \textbf{$\alpha$Transfer} \\
        \midrule
        Task Arithmetic \citep{ilharco2023editing} & Identity & Grid & Global & \cmark \\
        TIES \citep{yadav2023ties} & Trim + sign election & Grid & Global & \cmark \\
        DARE \citep{yu2024language} & Random drop + rescale & Grid & Global & \cmark \\
        TSV-M \citep{gargiulo2025task} & Low-rank SVD + whiten & Grid & Global & \cmark \\
        Subspace Boosting \citep{skorobogat2025subspace} & SVD + clamping & Grid & Global & \cmark \\
        AdaMerging \citep{yang2024adamerging} & Identity & Gradient & Task/Layer & \cmark \\
        AdaMerging++ \citep{yang2024adamerging} & Trim + sign election & Gradient & Task/Layer & \cmark \\
        DivMerge \citep{touayouch2026divmerge} & Identity & Gradient & Task/Layer & \cmark \\
        \midrule
        WUDI-Merging \citep{cheng2025whoever} & Per-layer optimization & -- & -- & \xmark \\
        \bottomrule
    \end{tabular}}
\end{table}

\subsection{The $\alpha$Transfer mechanism}
\label{ssec:coefficient_transfer}

We propose $\alpha$Transfer, a simple yet efficient approach for obtaining merging coefficients for large target models by leveraging small proxy models from the same family. The core insight driving this approach is that merging coefficients encode the relative importance of tasks and their interactions, which are primarily determined by the task characteristics rather than model scale. Consequently, coefficients learned on a small proxy model can be transferred to larger target models within the same family, significantly reducing the computational bottleneck of coefficient search.

\paragraph{Global and task-wise coefficient transfer.}
Given a target model $\theta_0^t$ and a proxy model $\theta_0^p$ from the same family, along with their respective fine-tuned models $\{\theta_i^t\}_{i=1}^T$ and $\{\theta_i^p\}_{i=1}^T$, coefficient transfer proceeds in two steps. First, we apply any existing merging method to the proxy model to obtain the merging coefficients:
\begin{equation}
    \boldsymbol{\alpha}^* = \arg\min_{\boldsymbol{\alpha}} \mathcal{L}^p\left(\theta_0^p + \sum_{i=1}^T \alpha_i \cdot \tilde{\tau}_i^p\right),
\end{equation}
where $\tilde{\tau}_i^p = f(\tau_i^p)$ are the edited task vectors with $\tau_i^p = \theta_i^p - \theta_0^p$, and $\mathcal{L}^p$ denotes the objective used by the chosen merging method on the proxy. For global methods, all $\alpha_i$ are tied to a single scalar $\alpha$. Note that the shape of $\boldsymbol{\alpha}^*$ depends only on the number of tasks $T$ (or is a single scalar for global methods), not on model size. We can therefore directly apply this same $\boldsymbol{\alpha}^*$ to merge the target model:
\begin{equation}
    \theta_{\text{merged}}^t = \theta_0^t + \sum_{i=1}^T \alpha_i^* \cdot \tilde{\tau}_i^t,
\end{equation}
where $\tilde{\tau}_i^t = f(\tau_i^t)$ with $\tau_i^t = \theta_i^t - \theta_0^t$. Note that only the coefficients $\boldsymbol{\alpha}^*$ are transferred from the proxy; the editing function $f(\cdot)$ is still applied natively to the target model's task vectors.

\paragraph{Layer-wise coefficient transfer.}
\label{sec:layerwise}
In layer-wise merging, each parameter layer is assigned its own merging coefficient. Since proxy and target models often differ in depth (and thus total number of layers), a direct one-to-one coefficient transfer is not always possible.

Modern architectures such as transformers consist of repeating blocks, where certain modules appear in every block (e.g., attention, feed-forward) while others do not (e.g., embeddings, output layers). For layers that do not repeat across blocks, coefficients are directly transferred as they appear in the same positions in both models. For layers that repeat across blocks, recent work has shown that layers at similar relative depths follow a shared progression of representations across depth \citep{wolfram2025layers}, suggesting that their merging coefficients might also be similar. Based on this intuition, we propose two strategies for transferring these coefficients.

\textbf{Copy-based transfer.} For a layer in block $b \in \{1, \ldots, B_t\}$ of the target model, we directly copy the coefficient from the corresponding layer type in the proxy block at the matched relative depth:
\begin{equation}
    b_{\text{proxy}} = \left\lceil \frac{b}{B_t} \cdot B_p \right\rceil,
\end{equation}
where $B_p$ and $B_t$ are the number of blocks in the proxy and target models. The ceiling ensures that target blocks 1 and $B_t$ map to proxy blocks 1 and $B_p$, respectively.

\textbf{Interpolation-based transfer.} To obtain smoother coefficient transitions, we linearly interpolate between the coefficients from the two nearest proxy blocks based on the target block's continuous relative position.

\section{Experiments}
\label{sec:experiments}

\begin{wraptable}{r}{0.36\linewidth}
\centering
\small
\caption{Models and parameter counts. For vision models, we report the image encoder only.}
\label{tab:model_param}
\begin{tabular}{llr}
\toprule
Family & Model & \#Params \\
\midrule
\multirow{3}{*}{CLIP-ViT} & ViT-B/32 & 88M \\
                          & ViT-B/16 & 86M \\
                          & ViT-L/14 & 304M \\
\midrule
\multirow{2}{*}{SigLIP}   & Base-256  & 93M \\
                          & Large-256 & 316M \\
\midrule
\multirow{2}{*}{Qwen3}    & 0.6B-Base  & 596M \\
                          & 4B-Base    & 4.02B \\
\bottomrule
\end{tabular}
\end{wraptable}

\paragraph{Experimental Setup.} We evaluate $\alpha$Transfer across three model families: CLIP-ViT \citep{radford2021learning} and SigLIP \citep{zhai2023sigmoid} for vision, and Qwen3 \citep{yang2025qwen3} for language. The parameter counts of these models are listed in \cref{tab:model_param}. Our evaluation covers 8 vision tasks (DTD \citep{cimpoi2014describing}, EuroSAT \citep{helber2018introducing}, FER2013 \citep{goodfellow2013challenges}, Food101 \citep{bossard2014food}, GTSRB \citep{stallkamp2011german}, RESISC45 \citep{cheng2017remote}, Stanford Cars \citep{krause20133d}, SUN397 \citep{xiao2016sun}) and 4 language tasks (Usefulness Judge \citep{chen2026expected}, IFEval \citep{zhou2023instruction}, Banking77 \citep{loukas2023making}, DDXPlus \citep{fansi2022ddxplus}). For each task, we fine-tune the corresponding base model to obtain an expert checkpoint. All experiments were conducted on a single NVIDIA RTX 4000 SFF Ada GPU, except for the Qwen3-4B model, which required two GPUs. Training and implementation details are provided in \cref{app:training_details}.

\subsection{Empirical analysis of coefficient transferability}

\begin{figure}[t]
\centering
\begin{minipage}[c]{0.43\linewidth}
    \centering
    \includegraphics[width=0.9\linewidth]{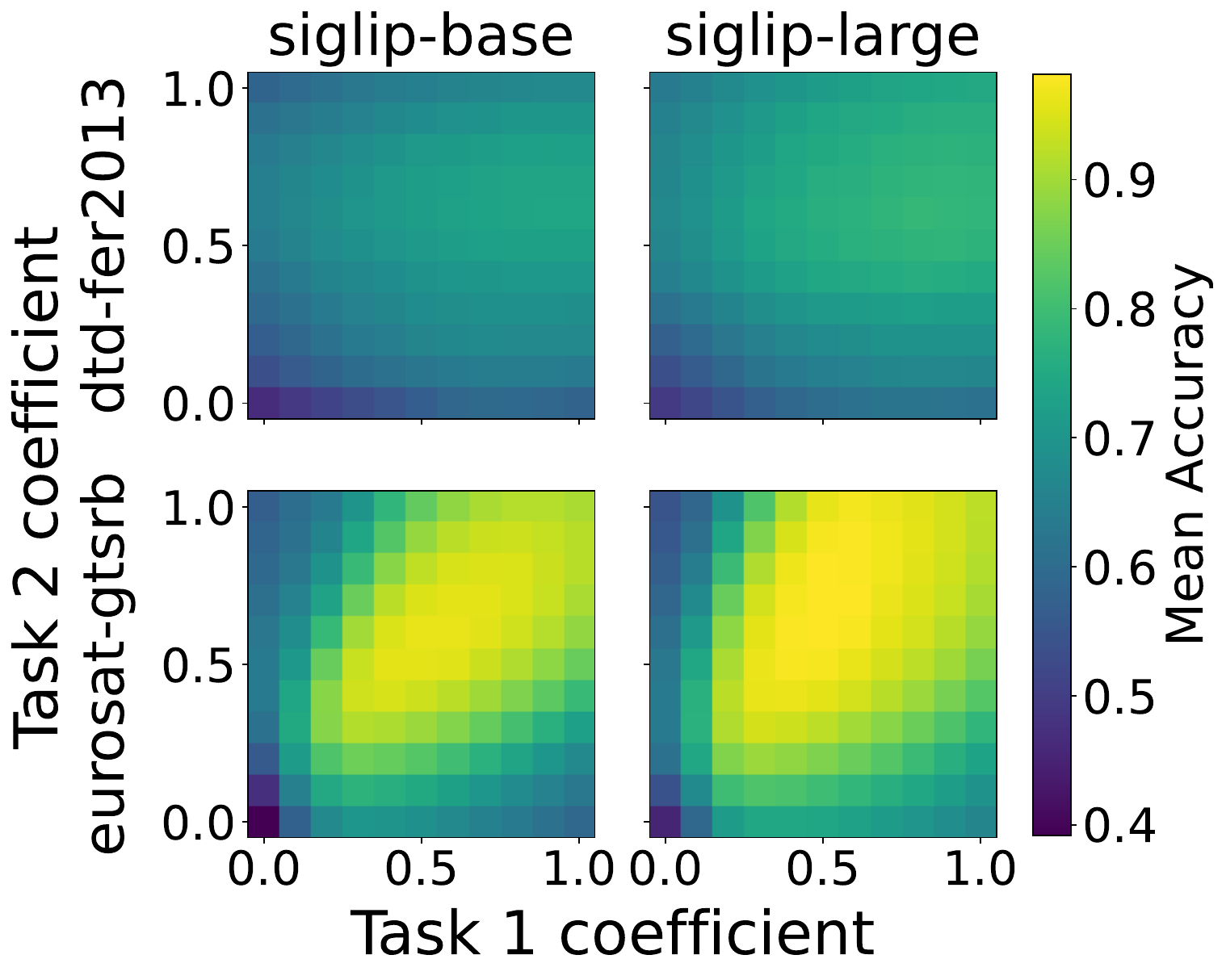}
    \captionof{figure}{\textbf{Performance landscapes across model scales.} We visualize mean accuracy over a grid of merging coefficients. The heatmaps show highly similar distributions, indicating that the overall shape of the performance landscape is preserved across different model sizes.}
    \label{fig:heatmap}
\end{minipage}
\hfill
\begin{minipage}[c]{0.52\linewidth}
    \centering
    \captionsetup{type=table}
    \caption{\textbf{Performance correlation between proxy and target models.} We sample multiple coefficient combinations, construct merged models accordingly, and compute the Spearman rank correlation ($\rho$) between their accuracies on the proxy and target models. The high correlations confirm that the relative performance of different coefficient combinations remains highly consistent across model sizes.}
    \label{tab:correlation}
    \small
    \begin{tabular}{llc}
        \toprule
        Proxy & Target & Spearman's $\rho$ \\
        \midrule
        ViT-B/32   & ViT-L/14   & 0.58 \\
        ViT-B/16   & ViT-L/14   & 0.84 \\
        SigLIP-B   & SigLIP-L   & 0.79 \\
        \midrule
        Qwen3-0.6B & Qwen3-1.7B & 0.94 \\
        Qwen3-0.6B & Qwen3-4B   & 0.90 \\
        \bottomrule
    \end{tabular}
\end{minipage}
\end{figure}

We first investigate whether the performance landscapes over the merging coefficient space exhibit structural similarity across model scales. This perspective is crucial: if the overall landscapes align, optimization algorithms will navigate comparable objective surfaces and converge to corresponding regions, making coefficient transfer viable even when the absolute global optimum is not found.

To build intuition, we visualize the performance landscape of merged models under different coefficient combinations. For a given pair of tasks, we vary the two corresponding coefficients over a grid and record the average accuracy. \cref{fig:heatmap} shows these landscapes for two representative task pairs on SigLIP-Base and SigLIP-Large. Despite the difference in model scale, the two models exhibit highly similar performance distributions over the coefficient space. This observation suggests that the overall topography of the coefficient space remains relatively stable across scales.

To rigorously verify this beyond specific task pairs, we extend our analysis to all three model families in a many-task setting. By sampling multiple coefficient combinations, we compute the Spearman rank correlation ($\rho$) between the proxy and target mean accuracies (averaged over the 8 vision or 4 language tasks, respectively). As detailed in \cref{tab:correlation}, all evaluated pairs exhibit consistently strong correlations ($\rho$ up to 0.94). This confirms that \textit{coefficient transferability} is a widespread and robust property in multi-task model merging, providing a solid empirical foundation for $\alpha$Transfer.\footnote{We also conduct experiments on cross-family model pairs with similar size. Results and discussion are provided in \cref{app:cross_family_transfer}.}

\subsection{Main results}
\label{sec:main_results}

\subsubsection{Vision models}
We evaluate $\alpha$Transfer across six merging methods\footnote{Results on two additional methods, TSV-M and Subspace Boosting, are provided in \cref{app:additional_methods}.} on the CLIP-ViT and SigLIP families across three proxy-target pairs: ViT-B/32 $\to$ ViT-L/14, ViT-B/16 $\to$ ViT-L/14, and SigLIP-Base $\to$ SigLIP-Large. For each method, we report the average accuracy under three settings: coefficients searched directly on the target model (origin), coefficients transferred from a proxy model (transfer), and simple averaging (SA) without any coefficient search, along with the computation time and peak memory cost of coefficient search.

As shown in \cref{tab:main_results}, $\alpha$Transfer achieves performance close to direct search across most methods and proxy-target pairs, while requiring only a fraction of the computational resources. In many cases the accuracy gap is negligible: on SigLIP-Base $\to$ SigLIP-Large, TIES and DARE drop by only 0.11\% and 0.32\%; on ViT-B/32 $\to$ ViT-L/14, Task Arithmetic and DARE drop by only 0.43\% and 0.49\%. Even in the most challenging cases (e.g., AdaMerging++ on ViT-B/32 $\to$ ViT-L/14, $-$7.02\%), the transferred model still surpasses the simple average baseline by 7.12\%, retaining most of the benefit of direct search.

\begin{table}[t]
\centering
\caption{
\textbf{Coefficient transfer on vision models.} We compare direct search (Origin), $\alpha$Transfer (Transfer) and simple average baseline (SA) across different proxy-target pairs. Transfer maintains accuracy comparable to the origin methods while significantly mitigating resource bottlenecks, yielding up to a 6.07$\times$ speedup and a 70.4\% memory reduction.}
\label{tab:main_results}

\setlength{\tabcolsep}{4pt}

\resizebox{\linewidth}{!}{
\begin{tabular}{clccccccccc}
\toprule

\multirow{2}{*}{\makecell{Proxy\\[-3pt]$\rightarrow$\\[-3pt]Target}}
& \multirow{2}{*}{Method}
& \multicolumn{3}{c}{Acc (\%)}
& \multicolumn{3}{c}{Time (h)}
& \multicolumn{3}{c}{Memory (GB)} \\

\cmidrule(lr){3-5}
\cmidrule(lr){6-8}
\cmidrule(lr){9-11}

&
& Origin & Transfer & SA
& Origin & Transfer & Speedup
& Origin & Transfer & Reduction \\

\midrule

\multirow{6}{*}{
\makecell[c]{
ViT-B/32 \\
$\rightarrow$ \\
ViT-L/14
}}
& Task Arithmetic
    & 75.98 & 75.55 & 72.89
    & 11.43 & 1.88 & 6.07$\times$
    & 2.12 & 0.67 & 68.7\% \\

& TIES
    & 75.17 & 71.65 & 64.96
    & 12.12 & 2.05 & 5.92$\times$
    & 2.12 & 0.67 & 68.7\% \\

& DARE
    & 76.05 & 75.56 & 62.70
    & 13.01 & 2.26 & 5.75$\times$
    & 2.12 & 0.67 & 68.7\% \\

& AdaMerging
    & 75.67 & 71.83 & 72.89
    & 1.69 & 0.72 & 2.35$\times$
    & 17.84 & 5.27 & 70.4\% \\

& AdaMerging++
    & 79.10 & 72.08 & 64.96
    & 1.46 & 0.83 & 1.77$\times$
    & 17.84 & 5.27 & 70.4\% \\

& DivMerge
    & 78.77 & 74.73 & 72.89
    & 0.33 & 0.15 & 2.14$\times$
    & 17.84 & 5.27 & 70.4\% \\

\midrule

\multirow{6}{*}{
\makecell[c]{
ViT-B/16 \\
$\rightarrow$ \\
ViT-L/14
}}
& Task Arithmetic
    & 75.98 & 72.89 & 72.89
    & 11.43 & 3.24 & 3.52$\times$
    & 2.12 & 0.87 & 58.8\% \\

& TIES
    & 75.17 & 69.53 & 64.96
    & 12.12 & 3.36 & 3.61$\times$
    & 2.12 & 0.87 & 58.8\% \\

& DARE
    & 76.05 & 72.91 & 62.70
    & 13.01 & 3.64 & 3.58$\times$
    & 2.12 & 0.87 & 58.8\% \\

& AdaMerging
    & 75.67 & 72.63 & 72.89
    & 1.69 & 0.94 & 1.79$\times$
    & 17.84 & 6.87 & 61.5\% \\

& AdaMerging++
    & 79.10 & 74.82 & 64.96
    & 1.46 & 0.84 & 1.75$\times$
    & 17.84 & 6.87 & 61.5\% \\

& DivMerge
    & 78.77 & 75.58 & 72.89
    & 0.33 & 0.13 & 2.52$\times$
    & 17.84 & 6.87 & 61.5\% \\

\midrule

\multirow{6}{*}{
\makecell[c]{
SigLIP-Base \\
$\rightarrow$ \\
SigLIP-Large
}}
& Task Arithmetic
    & 80.22 & 78.83 & 75.05
    & 10.40 & 3.46 & 3.01$\times$
    & 2.84 & 1.07 & 62.3\% \\

& TIES
    & 77.50 & 77.39 & 66.95
    & 11.75 & 3.90 & 3.01$\times$
    & 2.84 & 1.07 & 62.3\% \\

& DARE
    & 79.84 & 79.52 & 64.79
    & 13.41 & 4.00 & 3.35$\times$
    & 2.84 & 1.07 & 62.3\% \\

& AdaMerging
    & 81.48 & 80.27 & 75.05
    & 1.58 & 1.02 & 1.54$\times$
    & 18.12 & 7.41 & 59.1\% \\

& AdaMerging++
    & 81.88 & 77.83 & 66.95
    & 1.63 & 1.02 & 1.59$\times$
    & 18.12 & 7.41 & 59.1\% \\

& DivMerge
    & 82.98 & 80.01 & 75.05
    & 0.32 & 0.14 & 2.33$\times$
    & 18.12 & 7.41 & 59.1\% \\

\bottomrule
\end{tabular}
}
\end{table}

The efficiency gains reflect the distinct resource bottlenecks of each method type. For evaluation-based searching methods, the bottleneck is computation, where a single run already exceeds 10 hours on the target model, and $\alpha$Transfer reduces this by up to \textbf{6.07$\times$} (Task Arithmetic, ViT-B/32 $\to$ ViT-L/14), bringing the search time to under 2 hours. For gradient-based methods, the bottleneck is GPU memory: running optimization on the proxy model allows a batch size of 16, whereas the target model requires reducing to 4 to avoid out-of-memory errors. Coefficient transfer eliminates this constraint, reducing GPU memory consumption by up to \textbf{70.4\%} (AdaMerging/AdaMerging++, ViT-B/32 $\to$ ViT-L/14). Together, these results show that coefficient transfer enables efficient merging on large target models at a fraction of the cost, with minimal sacrifice in accuracy.

\subsubsection{Large language models}
To investigate coefficient transferability in LLMs, we evaluate Qwen3 models across two pairs: 0.6B $\to$ 1.7B and 0.6B $\to$ 4B. For TIES, we observe the DDXPlus expert is often filtered out due to conflicting update directions. To address this, we adopt a TIES-Pairwise variant: merging Usefulness Judge and IFEval in the first stage, then merging the result with Banking77 and DDXPlus in the second stage, using a consistent $\alpha$ throughout. This approach recovers DDXPlus performance while keeping the search space identical to standard TIES.

As shown in \cref{tab:llm_results}, $\alpha$Transfer yields performance that closely tracks the exhaustive search baseline. While the accuracy gap on the 1.7B model is within a marginal 2\%, $\alpha$Transfer achieves zero performance degradation for both Task Arithmetic and TIES-Pairwise on the 4B scale. More importantly, our method delivers a $20\times$ speedup and an 85\% memory reduction for the 4B model, compressing the memory requirement from over 40GB to under 6GB. These results demonstrate that $\alpha$Transfer effectively bypasses the prohibitive cost of autoregressive evaluation, making the merging of large-scale models both accessible and scalable.

\begin{table}[t]
\centering
\caption{
\textbf{Coefficient transfer on Qwen3 LLMs.} We compare direct search (Origin), coefficient transfer (Transfer) and simple average baseline (SA) across different proxy-target pairs. Transfer maintains accuracy comparable to the origin methods while significantly mitigating resource bottlenecks, yielding up to a 19.97$\times$ speedup and an 85.2\% memory reduction. We include a TIES-Pairwise variant to address the poor performance of the original TIES method.}
\label{tab:llm_results}
\setlength{\tabcolsep}{4pt}
\resizebox{\linewidth}{!}{
\begin{tabular}{clccccccccc}
\toprule
\multirow{2}{*}{Proxy $\rightarrow$ Target}
& \multirow{2}{*}{Method}
& \multicolumn{3}{c}{Acc (\%)}
& \multicolumn{3}{c}{Time (h)}
& \multicolumn{3}{c}{Memory (GB)} \\
\cmidrule(lr){3-5}
\cmidrule(lr){6-8}
\cmidrule(lr){9-11}
&
& Origin & Transfer & SA
& Origin & Transfer & Speedup
& Origin & Transfer & Reduction \\
\midrule
\multirow{3}{*}{0.6B $\rightarrow$ 1.7B}
& Task Arithmetic
    & 76.44 & 74.63 & 63.85
    & 0.78 & 0.07 & 11.00$\times$
    & 17.21 & 5.96 & 65.4\% \\
& TIES
    & 58.35 & 58.26 & 32.65
    & 1.86 & 0.10 & 18.96$\times$
    & 17.21 & 5.96 & 65.4\% \\
& TIES-Pairwise
    & 71.15 & 70.37 & 44.01
    & 1.55 & 0.08 & 19.94$\times$
    & 10.32 & 3.58 & 65.3\% \\
\midrule
\multirow{3}{*}{0.6B $\rightarrow$ 4B}
& Task Arithmetic
    & 76.03 & 76.03 & 71.91
    & 1.12 & 0.10 & 11.03$\times$
    & 40.22 & 5.96 & 85.2\% \\
& TIES
    & 66.04 & 65.26 & 41.04
    & 3.80 & 0.20 & 18.98$\times$
    & 40.22 & 5.96 & 85.2\% \\
& TIES-Pairwise
    & 73.69 & 73.69 & 59.57
    & 4.00 & 0.20 & 19.97$\times$
    & 24.13 & 3.58 & 85.2\% \\
\bottomrule
\end{tabular}
}
\end{table}

\subsection{Constrained budget analysis}
Beyond final accuracy, we examine whether coefficient transfer can match or exceed the performance of direct search given the same amount of computation spent on the target model.

\begin{figure}[t]
    \centering
    \begin{subfigure}[b]{0.32\linewidth}
        \includegraphics[width=\linewidth]{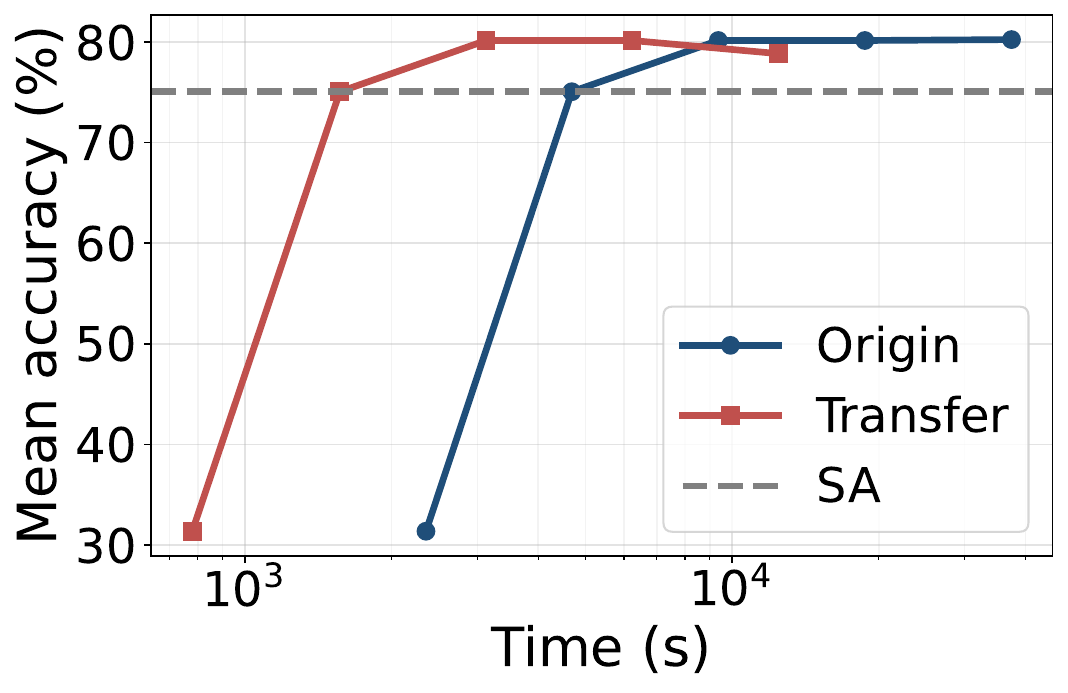}
        \caption{Task Arithmetic}
    \end{subfigure}
    \hfill
    \begin{subfigure}[b]{0.32\linewidth}
        \includegraphics[width=\linewidth]{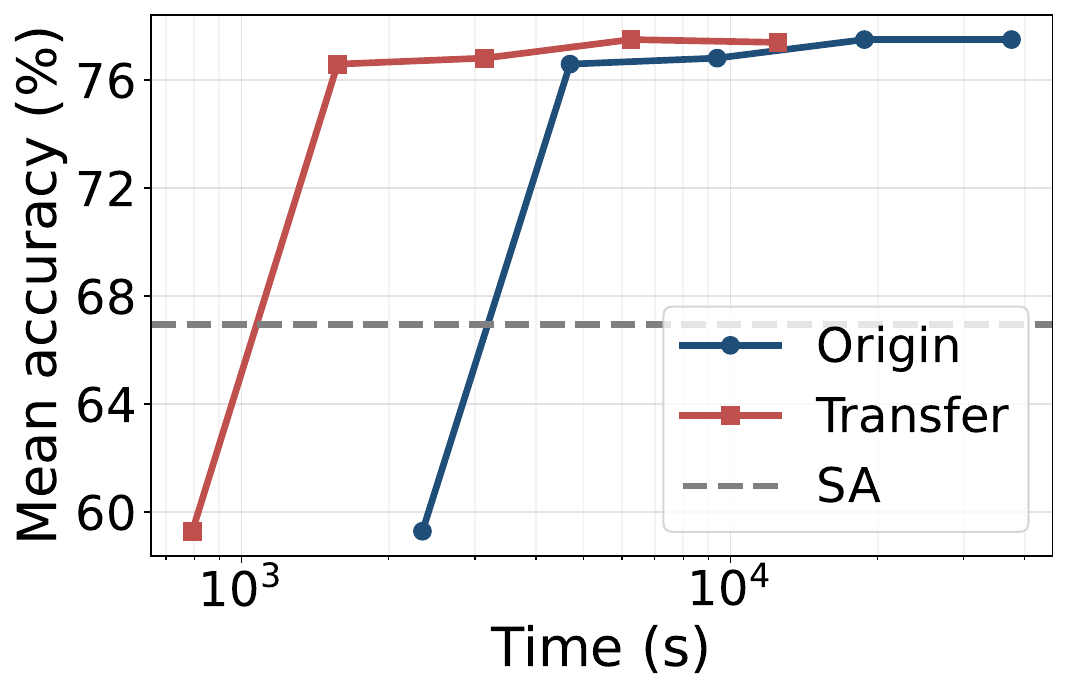}
        \caption{TIES}
    \end{subfigure}
    \hfill
    \begin{subfigure}[b]{0.32\linewidth}
        \includegraphics[width=\linewidth]{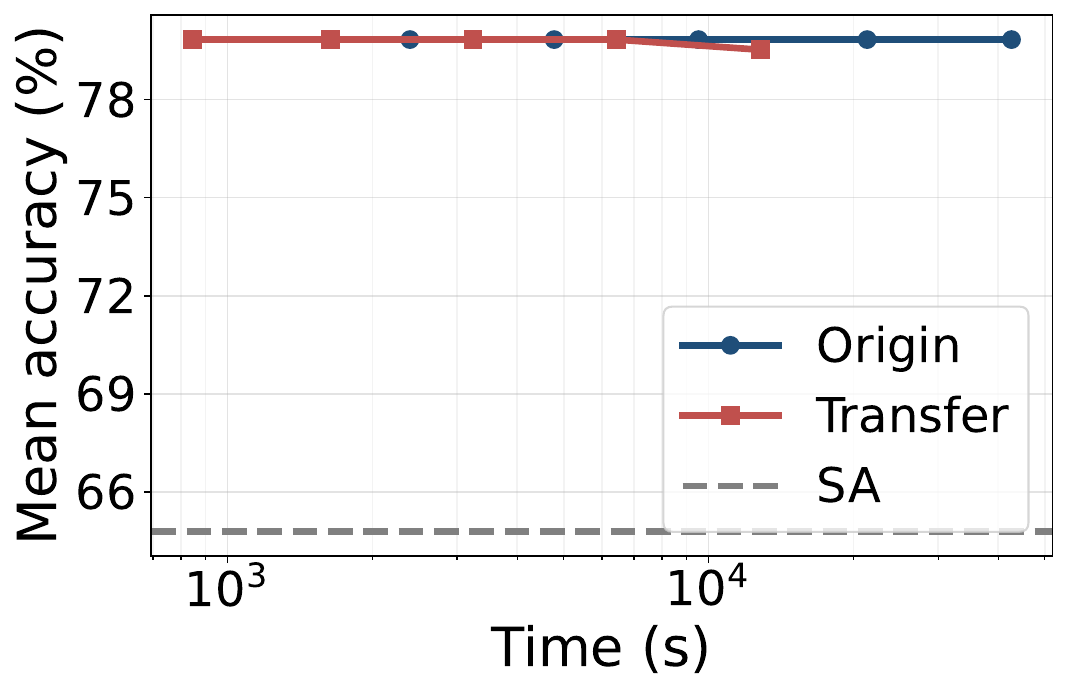}
        \caption{DARE}
    \end{subfigure}

    \vspace{0.5em}

    \begin{subfigure}[b]{0.32\linewidth}
        \includegraphics[width=\linewidth]{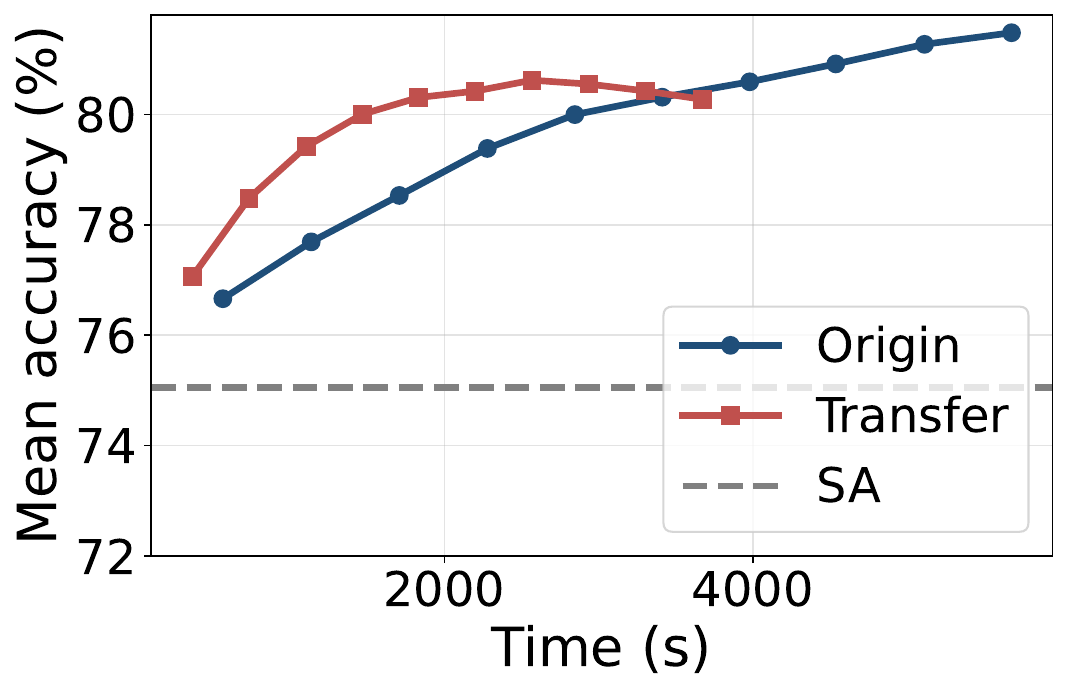}
        \caption{AdaMerging}
    \end{subfigure}
    \hfill
    \begin{subfigure}[b]{0.32\linewidth}
        \includegraphics[width=\linewidth]{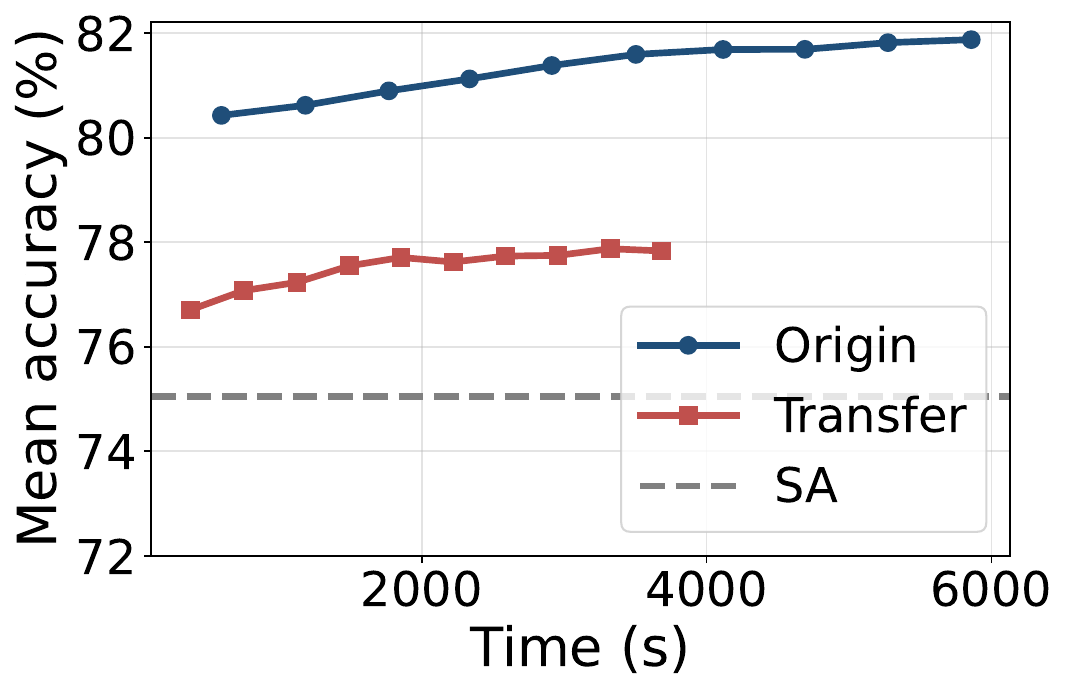}
        \caption{AdaMerging++}
    \end{subfigure}
    \hfill
    \begin{subfigure}[b]{0.32\linewidth}
        \includegraphics[width=\linewidth]{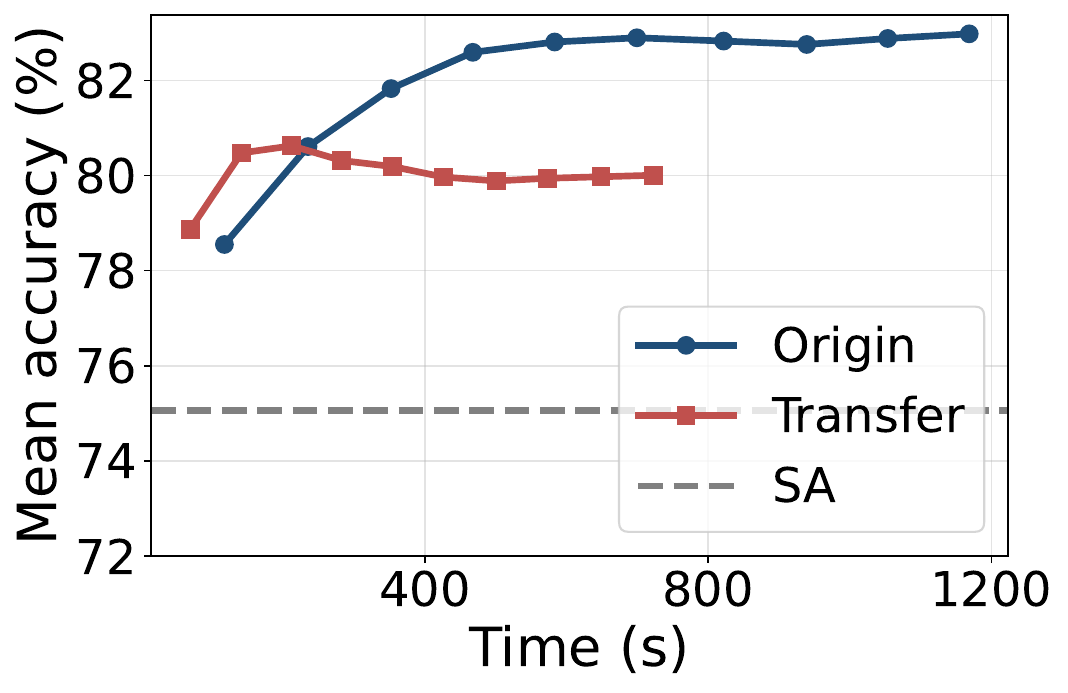}
        \caption{DivMerge}
    \end{subfigure}

    \caption{\textbf{Budget analysis on SigLIP models.} We compare the best achievable accuracy under varying computational time budgets for direct search (Origin) and $\alpha$Transfer (Transfer), with simple average (SA) as a static baseline. Transfer shifts the performance curves significantly to the left, reaching comparable accuracy in a fraction of the time for most methods. The performance gap in AdaMerging++ is further investigated in \cref{app:failure_cases}, with full results provided in \cref{app:efficiency_full_results}.}
    \label{fig:budget}
\end{figure}

\paragraph{Evaluation-based methods.}
We simulate increasing evaluation budgets by progressively refining the search granularity. At budget level $k$, we evaluate $2^k$ uniformly spaced points in the coefficient space ($k = 0, 1, \ldots, 4$), and report the best accuracy found so far. For transfer, we directly apply the proxy-optimal coefficient to the target model without any target-side evaluation. As shown in the top row of \cref{fig:budget}, transferred coefficients achieve strong performance even at low budgets across all three methods. For Task Arithmetic and TIES, the origin curve requires substantially more evaluations to reach comparable accuracy. For DARE, both curves remain nearly flat, with transfer matching origin performance and substantially outperforming the simple average.

\paragraph{Gradient-based methods.}
For gradient-based methods, we independently optimize coefficients on the proxy and target and evaluate intermediate proxy coefficients directly on the target. As shown in the bottom row of \cref{fig:budget}, $\alpha$Transfer for AdaMerging and DivMerge converges faster initially, achieving competitive or higher accuracy than direct target optimization under limited budgets. AdaMerging++ shows a performance gap due to overly high sparsity; reducing sparsity slightly lowers in-model performance but substantially improves transferability (\cref{app:failure_cases}).

\subsection{Layer-wise coefficient transfer}
We evaluate layer-wise coefficient transfer (detailed in \cref{sec:layerwise}) on AdaMerging and AdaMerging++ across three proxy-target pairs (\cref{tab:layerwise_results}). Copy-based and interpolation mapping achieve comparable performance, with an accuracy gap consistently below 0.5\%. On ViT-B/16 $\to$ ViT-L/14 and SigLIP-Base $\to$ SigLIP-Large, transfer matches or outperforms direct target optimization, while reducing computation time by 1.53$\times$--1.71$\times$ and memory usage by 59.1\%--61.5\%.

The only exception is ViT-B/32 $\to$ ViT-L/14, where transfer trails direct search by 1--3\%, consistent with its lower coefficient correlation (\cref{tab:correlation}, $\rho=0.58$ vs. 0.79--0.84). We hypothesize that the larger patch-size difference alters the optimization landscape, making direct coefficient transfer less effective.

\begin{table}[t]
\centering
\caption{\textbf{Layer-wise coefficient transfer.} We compare direct search (Origin) with two relative-depth mapping variants, Copy and Interpolation (Interp). With ViT-B/16 and SigLIP-Base as proxies, layer-wise transfer achieves higher accuracy than direct search on the target while reducing time and memory.}
\label{tab:layerwise_results}
\setlength{\tabcolsep}{4pt}
% \resizebox{\linewidth}{!}{
\scalebox{0.9}{
\begin{tabular}{clccccc}
\toprule
\multirow{2}{*}{Proxy $\rightarrow$ Target}
& \multirow{2}{*}{Method}
& \multicolumn{3}{c}{Acc (\%)}
& \multirow{2}{*}{\makecell{Time\\Speedup}}
& \multirow{2}{*}{\makecell{Memory\\Reduction}} \\
\cmidrule(lr){3-5}
&
& Origin & Copy & Interp & & \\
\midrule
\multirow{2}{*}{ViT-B/32 $\rightarrow$ ViT-L/14}
& AdaMerging   & \textbf{80.57} & 77.85 & 78.21 & 2.17$\times$ & 70.4\% \\
& AdaMerging++ & \textbf{82.80} & 81.35 & 81.55 & 2.17$\times$ & 70.4\% \\
\midrule
\multirow{2}{*}{ViT-B/16 $\rightarrow$ ViT-L/14}
& AdaMerging   & 80.57 & \textbf{81.21} & 81.13 & 1.68$\times$ & 61.5\% \\
& AdaMerging++ & 82.80 & \textbf{83.15} & 83.07 & 1.71$\times$ & 61.5\% \\
\midrule
\multirow{2}{*}{SigLIP-Base $\rightarrow$ SigLIP-Large}
& AdaMerging   & 83.57 & \textbf{84.25} & 84.21 & 1.53$\times$ & 59.1\% \\
& AdaMerging++ & 83.13 & 83.55 & \textbf{83.57} & 1.61$\times$ & 59.1\% \\
\bottomrule
\end{tabular}
}
\end{table}

\begin{table}[t]
\centering
\caption{\textbf{Hybrid search on vision models.} We compare direct search (Origin), coefficient transfer (Transfer), and hybrid search (Hybrid), which refines the transferred coefficients on the target model. Hybrid search recovers most of the accuracy lost by transfer while remaining faster than direct search.}
\label{tab:hybrid}
\setlength{\tabcolsep}{4pt}
% \resizebox{\linewidth}{!}{
\scalebox{0.9}{
\begin{tabular}{clccccc}
\toprule
\multirow{2}{*}{Proxy $\rightarrow$ Target}
& \multirow{2}{*}{Method}
& \multicolumn{3}{c}{Acc (\%)}
& \multicolumn{2}{c}{Speedup} \\
\cmidrule(lr){3-5}
\cmidrule(lr){6-7}
&
& Origin & Transfer & Hybrid
& Transfer & Hybrid \\
\midrule
\multirow{3}{*}{
\makecell[c]{
ViT-B/32 $\rightarrow$ ViT-L/14
}}
& Task Arithmetic
    & \textbf{75.98} & 75.55 & \textbf{75.98}
    & 6.07$\times$ & 1.86$\times$ \\
& AdaMerging
    & 75.67 & 71.83 & \textbf{76.00}
    & 2.35$\times$ & 1.40$\times$ \\
& DivMerge
    & \textbf{78.77} & 74.73 & 78.51
    & 2.14$\times$ & 1.38$\times$ \\
\midrule
\multirow{3}{*}{
\makecell[c]{
ViT-B/16 $\rightarrow$ ViT-L/14
}}
& Task Arithmetic
    & \textbf{75.98} & 72.89 & \textbf{75.98}
    & 3.52$\times$ & 1.68$\times$ \\
& AdaMerging
    & 75.67 & 72.63 & \textbf{76.41}
    & 1.79$\times$ & 1.28$\times$ \\
& DivMerge
    & 78.77 & 75.58 & \textbf{78.78}
    & 2.52$\times$ & 1.29$\times$ \\
\midrule
\multirow{3}{*}{
\makecell[c]{
SigLIP-Base $\rightarrow$ SigLIP-Large
}}
& Task Arithmetic
    & \textbf{80.22} & 78.83 & 78.96
    & 3.01$\times$ & 1.46$\times$ \\
& AdaMerging
    & 81.48 & 80.27 & \textbf{81.59}
    & 1.54$\times$ & 1.21$\times$ \\
& DivMerge
    & \textbf{82.98} & 80.01 & 82.92
    & 2.33$\times$ & 1.20$\times$ \\
\bottomrule
\end{tabular}
}
\vspace{-10pt}
\end{table}

\subsection{Extending $\alpha$Transfer via hybrid search}
\label{sec:hybrid_search}

While $\alpha$Transfer substantially reduces the cost of coefficient search, some applications may prioritize accuracy over efficiency. For these cases, we consider a hybrid extension that uses the transferred coefficients as an initialization and allocates a small budget on the target for refinement. For evaluation-based methods, we conduct a coarse search on the proxy followed by a local refinement on the target. For gradient-based methods, we perform half of the optimization steps on the proxy and complete the remaining steps on the target, starting from the transferred coefficients.

As shown in \cref{tab:hybrid}, hybrid search recovers most of the accuracy lost by direct transfer on vision models, matching or exceeding direct search in most settings with a 1.20$\times$ to 1.86$\times$ speedup. The same holds for LLMs (\cref{tab:hybrid_llm}): hybrid search recovers the accuracy of direct search for all proxy-target pairs, with 2.29$\times$ to 4.35$\times$ speedups. These results show that $\alpha$Transfer can be used either as a standalone low-cost method or as an initialization that accelerates full-scale search, allowing practitioners to trade off efficiency and accuracy.

\begin{table}[t]
\centering
\caption{
\textbf{Hybrid search on LLMs.} Hybrid search matches the accuracy of direct search (Origin) on all proxy-target pairs with a 2.29$\times$ to 4.35$\times$ speedup.
}
\label{tab:hybrid_llm}
\setlength{\tabcolsep}{4pt}
\scalebox{0.9}{
\begin{tabular}{llccccc}
\toprule
\multirow{2}{*}{Proxy $\rightarrow$ Target}
& \multirow{2}{*}{Method}
& \multicolumn{3}{c}{Acc (\%)}
& \multicolumn{2}{c}{Speedup} \\
\cmidrule(lr){3-5}
\cmidrule(lr){6-7}
&
& Origin & Transfer & Hybrid
& Transfer & Hybrid \\
\midrule
\multirow{3}{*}{Qwen3-0.6B $\rightarrow$ Qwen3-1.7B}
& Task Arithmetic
    & \textbf{76.44} & 74.63 & \textbf{76.44}
    & 11.00$\times$ & 2.29$\times$ \\
& TIES
    & \textbf{58.35} & 58.26 & \textbf{58.35}
    & 18.96$\times$ & 4.34$\times$ \\
& TIES-Pairwise
    & \textbf{71.15} & 70.37 & \textbf{71.15}
    & 19.94$\times$ & 3.05$\times$ \\
\midrule
\multirow{3}{*}{Qwen3-0.6B $\rightarrow$ Qwen3-4B}
& Task Arithmetic
    & \textbf{76.03} & \textbf{76.03} & \textbf{76.03}
    & 11.03$\times$ & 2.30$\times$ \\
& TIES
    & \textbf{66.04} & 65.26 & \textbf{66.04}
    & 18.98$\times$ & 4.35$\times$ \\
& TIES-Pairwise
    & \textbf{73.69} & \textbf{73.69} & \textbf{73.69}
    & 19.97$\times$ & 3.06$\times$ \\
\bottomrule
\end{tabular}
}
\end{table}

\section{Conclusion}
We introduced $\alpha$Transfer, a highly efficient framework that decouples the cost of coefficient search from target model size. By optimizing coefficients on a small proxy and transferring them to the target model, we achieve accuracy comparable to exhaustive direct search. Across six merging methods, this approach yields a $6\times$ speedup and 70\% memory reduction on vision transformers, and up to a $20\times$ speedup and 85\% memory reduction on large language models.

These findings suggest that optimal merging coefficients are primarily properties of the tasks and model family, rather than scale. This insight opens exciting directions for future work, such as investigating the transferability of other merging hyperparameters and exploring cross-recipe or cross-architecture coefficient mapping. Ultimately, we hope $\alpha$Transfer serves as a crucial step toward making model merging practical and highly scalable for modern foundation models.

\clearpage

\section*{Acknowledgements}
This work was supported in part by the National Science and Technology Council, Taiwan, under Grants 114-2628-E-002-021-, 114-2628-E-A49-002, 115-2634-F-002-012-, 115-2223-E-002-005-MY3, and 115-2218-E-002-026-, and the Taiwan Centers of Excellence in Artificial Intelligence. Shao-Hua Sun was supported by the Yushan Fellow Program of the Ministry of Education, Taiwan.

% \section*{AI use statement}
% \input{sections/ai_use_statement}

% \section*{Reproducibility statement}
% \input{sections/reproducibility_statement}

\bibliography{iclr2027_conference}
\bibliographystyle{iclr2027_conference}

\clearpage

\appendix
\crefalias{section}{appendix}
\crefalias{subsection}{appendix}
\crefalias{subsubsection}{appendix}

\section*{Appendix}

\vspace{-1.8cm}
\begingroup
\hypersetup{colorlinks=false, linkcolor=black}
\hypersetup{pdfborder={0 0 0}}
\part{} % Start the appendix part
\parttoc % Insert the appendix TOC
\endgroup

% 1. Tell the TOC to start tracking sections again
% \addtocontents{toc}{\protect\setcounter{tocdepth}{2}}

% % 2. Style the TOC header
% \etocsettocstyle{\subsection*{Contents of Appendix}}{}

% % 3. Print the standard global TOC (which is now filtered)
% \begingroup
% \hypersetup{hidelinks}
% \tableofcontents % Or whatever custom command you are using for the Appendix TOC
% \endgroup
% \clearpage

\section{Training details}
\label{app:training_details}

In this section, we provide the training details for the models used in our experiments. We separate the discussion into vision models (\cref{app:vision_models}) and large language models (\cref{app:llms}).

\subsection{Vision models}
\label{app:vision_models}

We adopt the pre-trained CLIP vision encoders released by OpenAI, specifically ViT-B/32, ViT-B/16, and ViT-L/14, as well as the SigLIP vision encoders released by Google (SigLIP-Base/16-256 and SigLIP-Large/16-256). Each model is fine-tuned independently on eight downstream image classification tasks: DTD, EuroSAT, FER2013, Food101, GTSRB, RESISC45, Stanford Cars, and SUN397.

\paragraph{Hyperparameter search.}
For each (model, dataset) pair, we perform a grid search over the following hyperparameters:
\begin{itemize}
    \item Batch size: $\{32, 48\}$
    \item Number of epochs: $\{5, 8, 10, 20\}$
    \item Learning rate: $\{1\mathrm{e}{-5},\ 3\mathrm{e}{-5}\}$
\end{itemize}
We train each configuration for up to the maximum number of epochs specified in the search space, evaluate the model on the validation set at the end of every epoch, and select the checkpoint that achieves the highest validation accuracy. The chosen hyperparameters used for each (model, dataset) combination are reported in \cref{tab:vision_hparams}. Notably, ViT-B/32 and ViT-L/14 share nearly identical optimal hyperparameters across all tasks (differing only on GTSRB), while the optimal configurations for ViT-B/16 and the SigLIP variants vary more substantially across datasets.

\paragraph{Optimization.}
All vision models are fine-tuned with the AdamW optimizer ($\beta_1 = 0.9$, $\beta_2 = 0.999$) and a linear learning-rate scheduler with a warmup ratio of $0.1$.

\begin{table}[h]
    \centering
    \caption{Selected hyperparameters for each vision model and dataset. Each cell reports the number of epochs (E), batch size (B), and learning rate (LR) chosen by the validation-accuracy criterion described in \cref{app:vision_models}.}
    \label{tab:vision_hparams}
    \small
    \setlength{\tabcolsep}{3pt}
    \begin{tabular}{l ccc ccc ccc ccc ccc}
        \toprule
        & \multicolumn{3}{c}{\textbf{ViT-B/32}} & \multicolumn{3}{c}{\textbf{ViT-B/16}} & \multicolumn{3}{c}{\textbf{ViT-L/14}} & \multicolumn{3}{c}{\textbf{SigLIP-B}} & \multicolumn{3}{c}{\textbf{SigLIP-L}} \\
        \cmidrule(lr){2-4} \cmidrule(lr){5-7} \cmidrule(lr){8-10} \cmidrule(lr){11-13} \cmidrule(lr){14-16}
        \textbf{Dataset} & E & B & LR & E & B & LR & E & B & LR & E & B & LR & E & B & LR \\
        \midrule
        DTD            & 10 & 32 & 1e-5 & 5  & 32 & 1e-5 & 10 & 32 & 1e-5 & 5  & 48 & 3e-5 & 5  & 48 & 3e-5 \\
        EuroSAT        & 10 & 32 & 1e-5 & 10 & 48 & 3e-5 & 10 & 32 & 1e-5 & 10 & 48 & 3e-5 & 10 & 48 & 3e-5 \\
        FER2013        & 10 & 32 & 1e-5 & 5  & 32 & 1e-5 & 10 & 32 & 1e-5 & 20 & 48 & 3e-5 & 20 & 48 & 3e-5 \\
        Food101        & 10 & 32 & 1e-5 & 5  & 32 & 1e-5 & 10 & 32 & 1e-5 & 20 & 48 & 3e-5 & 10 & 48 & 1e-5 \\
        GTSRB          & 10 & 48 & 3e-5 & 5  & 32 & 1e-5 & 10 & 32 & 1e-5 & 5  & 48 & 3e-5 & 5  & 48 & 3e-5 \\
        RESISC45       & 10 & 32 & 1e-5 & 8  & 32 & 3e-5 & 10 & 32 & 1e-5 & 5  & 48 & 3e-5 & 10 & 48 & 1e-5 \\
        Stanford Cars  & 10 & 32 & 1e-5 & 10 & 48 & 3e-5 & 10 & 32 & 1e-5 & 20 & 48 & 3e-5 & 10 & 48 & 1e-5 \\
        SUN397         & 10 & 32 & 1e-5 & 5  & 32 & 1e-5 & 10 & 32 & 1e-5 & 20 & 48 & 3e-5 & 10 & 48 & 1e-5 \\
        \bottomrule
    \end{tabular}
\end{table}

\subsection{Large language models}
\label{app:llms}

We fine-tune Qwen3 models at three scales (0.6B, 1.7B, and 4B parameters) on four downstream tasks: Banking77, DDXPlus, instruction following, and usefulness judging.

\paragraph{Training data.}
For Banking77 and DDXPlus, we fine-tune directly on the training split of the corresponding dataset. For instruction following, we train on a $15{,}000$-example subsample drawn from the Nemotron-SFT-Instruction-Following-Chat-v2 dataset\footnote{\url{https://huggingface.co/datasets/nvidia/Nemotron-SFT-Instruction-Following-Chat-v2}}. For the usefulness-judging task, we train on the usefulness-judge dataset\footnote{\url{https://huggingface.co/datasets/miulab/usefulness-judge}} proposed by \citet{chen2026expected}.

\paragraph{Hyperparameter search.}
For each (model, task) pair, we perform a grid search over batch sizes $\{36, 48\}$ and learning rates $\{1\mathrm{e}{-5},\ 4\mathrm{e}{-5}\}$. We train each configuration for up to $6$ epochs, evaluate on the validation set at the end of every epoch, and select the checkpoint with the highest validation accuracy. The selected configuration is the same across all (model, task) pairs: batch size $36$ and learning rate $4\mathrm{e}{-5}$.

\paragraph{Optimization.}
All LLMs are fully fine-tuned using the AdamW optimizer with a cosine learning-rate scheduler.

\clearpage

\section{Evaluation details}
\label{app:evaluation_details}

\subsection{Model merging hyperparameters}

For optimization-based methods (AdaMerging, AdaMerging++, and DivMerge), we dynamically adjust the batch size based on the model scale to accommodate memory constraints: we use a batch size of 16 for base-scale models (SigLIP-Base, ViT-B/32, ViT-B/16) and a batch size of 4 for large-scale models (SigLIP-Large, ViT-L/14).

\paragraph{Search Range for Coefficients.} For evaluation-based methods that require a grid search over the merging coefficient $\alpha$ (as reported in \cref{tab:main_results}), we define the search space following the recommendations in their respective original works:
\begin{itemize}
    \item \textbf{Task Arithmetic:} We search within the range $(0, 1]$ using 16 equidistant points (an interval of 0.0625).
    \item \textbf{DARE and TIES:} We search within the range $(0, 1.8]$ using 18 points (an interval of 0.1).
\end{itemize}

\paragraph{Method-specific Settings.} The structural and optimization hyperparameters for each method are as follows:
\begin{itemize}
    \item \textbf{AdaMerging:} We optimize coefficients for 500 steps using the Adam optimizer with a learning rate of $1\times 10^{-3}$. We use 1,000 samples to compute the merging objective and set the initial coefficient to 0.5.
    
    \item \textbf{AdaMerging++:} We adopt the identical optimization settings as AdaMerging. Additionally, we incorporate TIES-style masking with a top-$k$ retention ratio of 20\% ($k=20$) and use summation as the merging function.
    
    \item \textbf{DivMerge:} The coefficients are optimized for 100 steps using the Adam optimizer with a learning rate of $1\times 10^{-2}$. We set the initial coefficient to 0.5, use 200 samples for the optimization process, and employ the Jensen-Shannon (JS) divergence as the metric.
    
    \item \textbf{TIES:} We apply a top-$k$ filtering ratio of 20\% ($k=20$) and use summation to merge the weight updates.
    
    \item \textbf{DARE:} We apply a random drop rate of 0.5 to the model weights before merging.
\end{itemize}

\clearpage
\section{Cross-family coefficient transferability}
\label{app:cross_family_transfer}

To examine whether coefficient transferability is specific to a model family, we evaluate size-matched cross-family pairs between CLIP-ViT and SigLIP using the same set of sampled coefficient combinations as in \cref{tab:correlation}. Results are shown in \cref{tab:cross_family_correlation}.

\begin{table}[h]
\centering
\caption{Spearman rank correlation ($\rho$) of mean accuracy between model pairs over the same set of sampled merging coefficients. Cross-family pairs are size-matched between CLIP-ViT and SigLIP.}
\label{tab:cross_family_correlation}
\begin{tabular}{llc}
\toprule
Group & Model Pair & Spearman $\rho$ \\
\midrule
\multirow{3}{*}{Same family}
    & ViT-B/16 -- ViT-L/14 & 0.84 \\
    & ViT-B/32 -- ViT-L/14 & 0.58 \\
    & SigLIP-Base -- SigLIP-Large & 0.79 \\
\midrule
\multirow{3}{*}{Cross family} 
    & ViT-B/16 -- SigLIP-Base & 0.83 \\
    & ViT-B/32 -- SigLIP-Base & 0.56 \\
    & ViT-L/14 -- SigLIP-Large & 0.74 \\
\bottomrule
\end{tabular}
\end{table}

Cross-family correlations fall in a similar range to same-family ones ($\rho = 0.56$ to $0.83$ vs.\ $0.58$ to $0.84$). We attribute this to two factors. First, all pairs are evaluated on an identical set of coefficients, and some task-level properties affect all models similarly; for instance, configurations with a large coefficient sum or a single dominant task coefficient tend to perform poorly regardless of the model. Second, both families share ViT backbones trained with contrastive image-text objectives, which may further contribute to the similarity. We also observe that the lowest correlations in both groups involve ViT-B/32, suggesting that architectural details such as patch size may matter more than the model family.

These results remain consistent with our main claim, as coefficient rankings are well correlated within each family. A systematic study of cross-family transfer, including end-to-end merging performance, is left to future work.

\clearpage
\section{Effect of Fine-Tuning Hyperparameters}
\label{app:finetune_recipe}

To analyze how the fine-tuning recipe of proxy experts affects transfer, we fix the target experts (ViT-L/14, fine-tuned with learning rate 1e-5 for 10 epochs) and vary the training recipe of the proxy experts (ViT-B/32 and ViT-B/16). For Task Arithmetic, the coefficient is searched over $\{0.1, 0.2, \dots, 1.0\}$, so the results may differ slightly from those in the main paper. For AdaMerging on experts fine-tuned with learning rate 3e-5, we run 1,500 optimization steps instead of 500 for both proxy and target, as AdaMerging on the ViT-L/14 experts does not converge within 500 steps under this setting.

\paragraph{Training duration.}
We fine-tune the proxy experts with learning rate 1e-5 for 10 epochs and use the intermediate checkpoints at epochs 1, 3, 5, and 10. As shown in \cref{tab:proxy_epochs}, Task Arithmetic transfers without loss in all cases, and AdaMerging stays within 1 point of direct search once the proxy experts are trained for 3 epochs or more. Only under-trained proxies (epoch 1) lead to noticeable drops for AdaMerging.

\begin{table}[h]
\centering
\caption{\textbf{Effect of proxy training duration.} Accuracy (\%) of direct search on the target (Origin) and of $\alpha$Transfer using proxy experts trained for different numbers of epochs. Target experts are fixed.}
\label{tab:proxy_epochs}
\setlength{\tabcolsep}{5pt}
\begin{tabular}{ll c cccc}
\toprule
\multirow{2}{*}{Proxy $\rightarrow$ Target}
& \multirow{2}{*}{Method}
& \multirow{2}{*}{Origin}
& \multicolumn{4}{c}{Transfer (proxy epochs)} \\
\cmidrule(lr){4-7}
& & & 1 & 3 & 5 & 10 \\
\midrule
\multirow{2}{*}{ViT-B/32 $\rightarrow$ ViT-L/14}
& Task Arithmetic & 75.99 & 75.99 & 75.99 & 75.99 & 75.99 \\
& AdaMerging      & 75.70 & 73.26 & 76.50 & 76.28 & 76.59 \\
\midrule
\multirow{2}{*}{ViT-B/16 $\rightarrow$ ViT-L/14}
& Task Arithmetic & 75.99 & 75.99 & 75.99 & 75.99 & 75.99 \\
& AdaMerging      & 75.70 & 70.62 & 75.46 & 75.90 & 76.13 \\
\bottomrule
\end{tabular}
\end{table}

\paragraph{Learning rate.}
We fine-tune the proxy experts for 10 epochs with learning rate 1e-5 or 3e-5. When the proxy is fine-tuned with a larger learning rate than the target, transfer performance drops (\cref{tab:proxy_lr}). For Task Arithmetic, the proxy selects $\alpha = 0.1$ instead of the target's optimal $\alpha = 0.2$, resulting in a 4.92-point drop.

To examine whether this drop is caused by the learning rate itself or by the mismatch between proxy and target, we further fine-tune the target experts with learning rate 3e-5, so that both share the same recipe. In this matched setting, all model sizes select $\alpha = 0.1$, and Task Arithmetic transfers without any loss. This indicates that the optimal coefficients depend on the fine-tuning recipe, but shift consistently across scales when the recipe is shared. For AdaMerging, matching the recipe reduces the drop on ViT-B/32 (from 2.53 to 1.51 points) but not on ViT-B/16 (2.95 vs.\ 3.03 points), suggesting that learning-based methods remain more stable when proxy and target both use a moderate learning rate.

\begin{table}[t]
\centering
\caption{\textbf{Effect of fine-tuning learning rate.} Accuracy (\%) of direct search on the target (Origin) and of $\alpha$Transfer under matched and mismatched learning rates between proxy and target experts.}
\label{tab:proxy_lr}
\setlength{\tabcolsep}{5pt}
\begin{tabular}{ll cc ccc}
\toprule
Proxy $\rightarrow$ Target
& Method
& \makecell{Proxy\\LR}
& \makecell{Target\\LR}
& Origin & Transfer & $\Delta$ \\
\midrule
\multirow{6}{*}{ViT-B/32 $\rightarrow$ ViT-L/14}
& \multirow{3}{*}{Task Arithmetic}
    & 1e-5 & 1e-5 & 75.99 & 75.99 & 0.00 \\
& & 3e-5 & 1e-5 & 75.99 & 71.07 & $-$4.92 \\
& & 3e-5 & 3e-5 & 68.90 & 68.90 & 0.00 \\
\cmidrule(lr){2-7}
& \multirow{3}{*}{AdaMerging}
    & 1e-5 & 1e-5 & 75.70 & 76.59 & $+$0.89 \\
& & 3e-5 & 1e-5 & 75.70 & 73.17 & $-$2.53 \\
& & 3e-5 & 3e-5 & 68.04 & 66.53 & $-$1.51 \\
\midrule
\multirow{6}{*}{ViT-B/16 $\rightarrow$ ViT-L/14}
& \multirow{3}{*}{Task Arithmetic}
    & 1e-5 & 1e-5 & 75.99 & 75.99 & 0.00 \\
& & 3e-5 & 1e-5 & 75.99 & 71.07 & $-$4.92 \\
& & 3e-5 & 3e-5 & 68.90 & 68.90 & 0.00 \\
\cmidrule(lr){2-7}
& \multirow{3}{*}{AdaMerging}
    & 1e-5 & 1e-5 & 75.70 & 76.13 & $+$0.43 \\
& & 3e-5 & 1e-5 & 75.70 & 72.75 & $-$2.95 \\
& & 3e-5 & 3e-5 & 68.04 & 65.01 & $-$3.03 \\
\bottomrule
\end{tabular}
\end{table}

\paragraph{Discussion.}
In our main experiments, the fine-tuning configuration of each checkpoint is selected independently (\cref{app:training_details}), reflecting realistic scenarios where checkpoints are fine-tuned separately. While most selected checkpoints use learning rate 1e-5, some use 3e-5, which may contribute to the accuracy drops observed in \cref{tab:main_results}. In practice, transfer is more reliable when the proxy experts are sufficiently trained and fine-tuned with the same recipe as the target.

\clearpage
\section{Results on Additional Merging Methods}
\label{app:additional_methods}

We further evaluate $\alpha$Transfer on two recent subspace-based merging methods, TSV-M \citep{gargiulo2025task} and Subspace Boosting \citep{skorobogat2025subspace}, the latter applied on top of Task Arithmetic (TA) and TIES-Merging (TIES). For all methods, merging coefficients are searched over $\{0.1, 0.2, \dots, 1.0\}$. For Subspace Boosting, we additionally search $\beta \in \{0, 0.01, 0.02\}$ on ViT-B/32 and fix the selected value $\beta = 0$, which equalizes all singular values for all other models, including direct search on the target. For Subspace Boosting (TIES), we follow the hyperparameter of TIES-Merging and keep the top 20\% of parameters.

As shown in \cref{tab:additional_methods}, $\alpha$Transfer achieves a $2.6\times$ to $5.6\times$ speedup and a 59\% to 69\% memory reduction across all settings. Accuracy is preserved within 1.35 points in all cases except Subspace Boosting (TIES) on ViT, where the drop is 2.57 points. In several cases, including Subspace Boosting (TA) and all SigLIP settings, the proxy selects the same coefficients as direct search on the target, yielding identical accuracy. These results indicate that $\alpha$Transfer generalizes to these additional methods.

\begin{table}[h]
\centering
\caption{
\textbf{Coefficient transfer on subspace-based merging methods.} We compare direct search (Origin), $\alpha$Transfer (Transfer) and simple average baseline (SA) on TSV-M and Subspace Boosting applied to Task Arithmetic (TA) and TIES-Merging (TIES).}
\label{tab:additional_methods}

\setlength{\tabcolsep}{4pt}

\resizebox{\linewidth}{!}{
\begin{tabular}{clccccccccc}
\toprule

\multirow{2}{*}{\makecell{Proxy\\[-3pt]$\rightarrow$\\[-3pt]Target}}
& \multirow{2}{*}{Method}
& \multicolumn{3}{c}{Acc (\%)}
& \multicolumn{3}{c}{Time (h)}
& \multicolumn{3}{c}{Memory (GB)} \\

\cmidrule(lr){3-5}
\cmidrule(lr){6-8}
\cmidrule(lr){9-11}

&
& Origin & Transfer & SA
& Origin & Transfer & Speedup
& Origin & Transfer & Reduction \\

\midrule

\multirow{3}{*}{
\makecell[c]{
ViT-B/32 \\
$\rightarrow$ \\
ViT-L/14
}}
& TSV-M
    & 86.29 & 85.74 & 68.58
    & 6.96 & 1.28 & 5.46$\times$
    & 2.12 & 0.66 & 68.7\% \\
& Subspace Boosting (TA)
    & 85.39 & 85.39 & 85.14
    & 6.89 & 1.29 & 5.33$\times$
    & 2.12 & 0.66 & 68.7\% \\
& Subspace Boosting (TIES)
    & 81.85 & 79.28 & 74.16
    & 6.94 & 1.25 & 5.55$\times$
    & 2.12 & 0.66 & 68.7\% \\

\midrule

\multirow{3}{*}{
\makecell[c]{
ViT-B/16 \\
$\rightarrow$ \\
ViT-L/14
}}
& TSV-M
    & 86.29 & 84.94 & 68.58
    & 6.96 & 1.85 & 3.77$\times$
    & 2.12 & 0.87 & 58.9\% \\
& Subspace Boosting (TA)
    & 85.39 & 85.39 & 85.14
    & 6.89 & 1.91 & 3.61$\times$
    & 2.12 & 0.87 & 58.9\% \\
& Subspace Boosting (TIES)
    & 81.85 & 79.28 & 74.16
    & 6.94 & 1.90 & 3.65$\times$
    & 2.12 & 0.87 & 58.9\% \\

\midrule

\multirow{3}{*}{
\makecell[c]{
SigLIP-Base \\
$\rightarrow$ \\
SigLIP-Large
}}
& TSV-M
    & 86.69 & 86.69 & 69.60
    & 6.14 & 2.33 & 2.63$\times$
    & 2.84 & 1.07 & 62.2\% \\
& Subspace Boosting (TA)
    & 84.87 & 84.87 & 85.20
    & 6.22 & 2.33 & 2.67$\times$
    & 2.84 & 1.07 & 62.2\% \\
& Subspace Boosting (TIES)
    & 80.02 & 80.02 & 74.17
    & 6.29 & 2.32 & 2.71$\times$
    & 2.84 & 1.07 & 62.2\% \\

\bottomrule
\end{tabular}
}
\end{table}

\clearpage
\section{Efficiency analysis complete results}
\label{app:efficiency_full_results}

\begin{figure}[h]
    \centering
    \begin{subfigure}[b]{0.32\linewidth}
        \includegraphics[width=\linewidth]{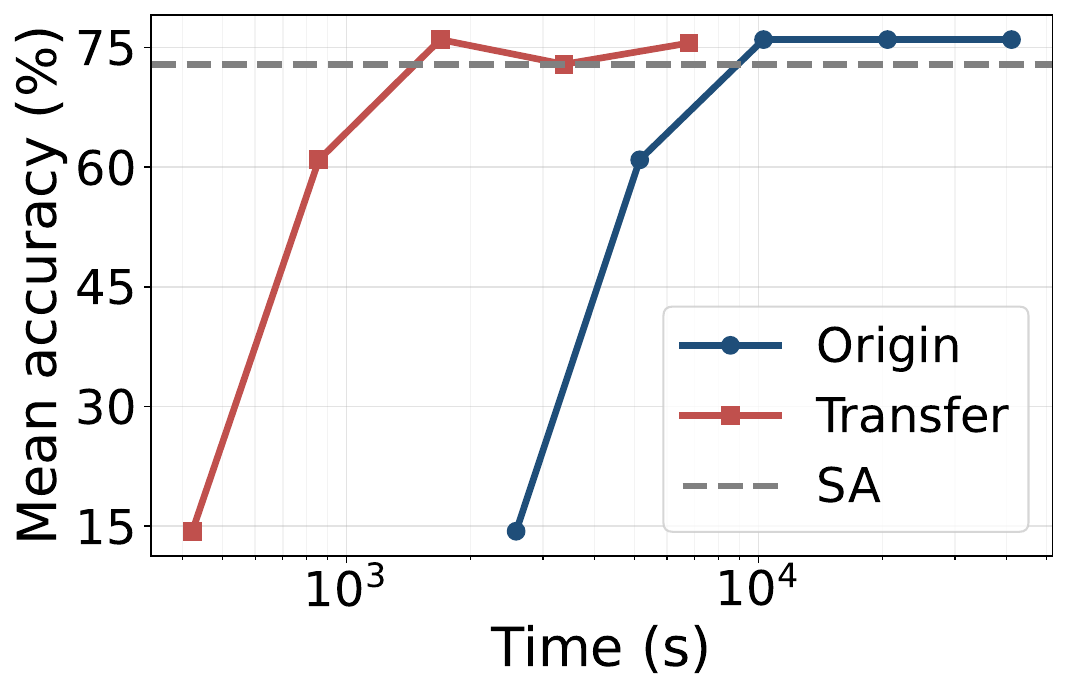}
        \caption{Task Arithmetic}
    \end{subfigure}
    \hfill
    \begin{subfigure}[b]{0.32\linewidth}
        \includegraphics[width=\linewidth]{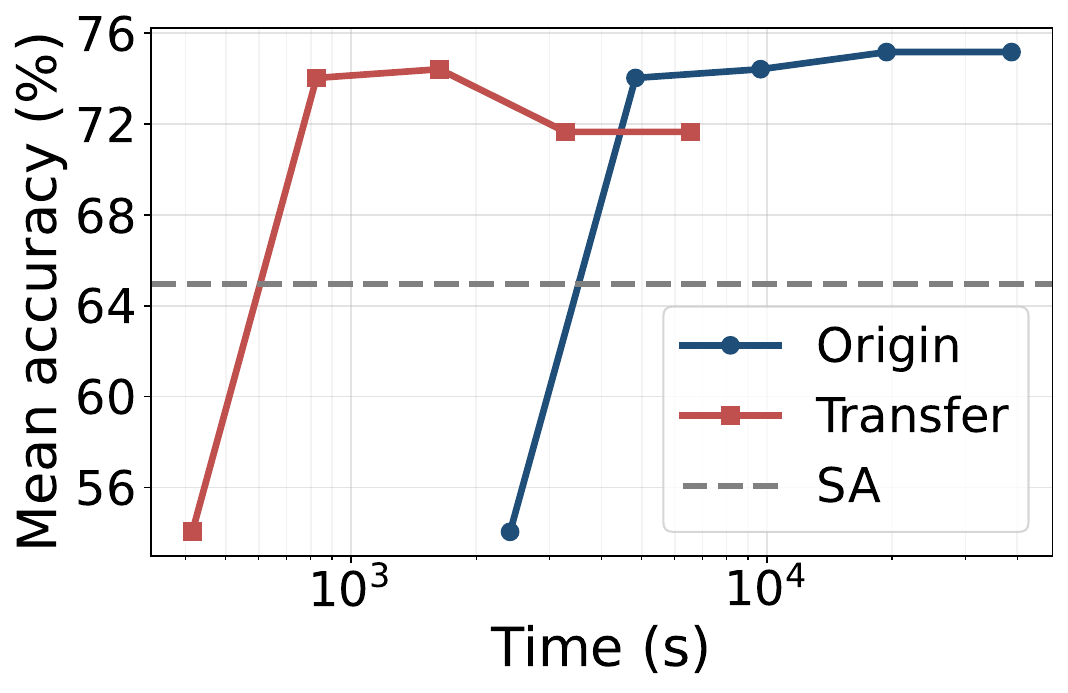}
        \caption{TIES}
    \end{subfigure}
    \hfill
    \begin{subfigure}[b]{0.32\linewidth}
        \includegraphics[width=\linewidth]{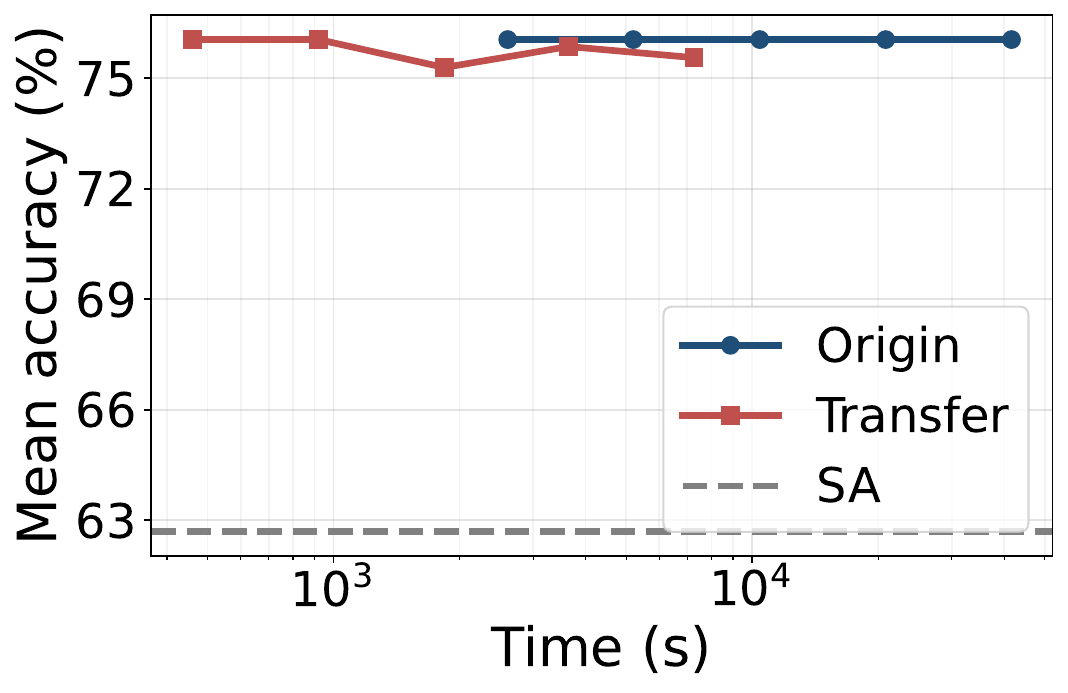}
        \caption{DARE}
    \end{subfigure}
    \vspace{0.5em}
    \begin{subfigure}[b]{0.32\linewidth}
        \includegraphics[width=\linewidth]{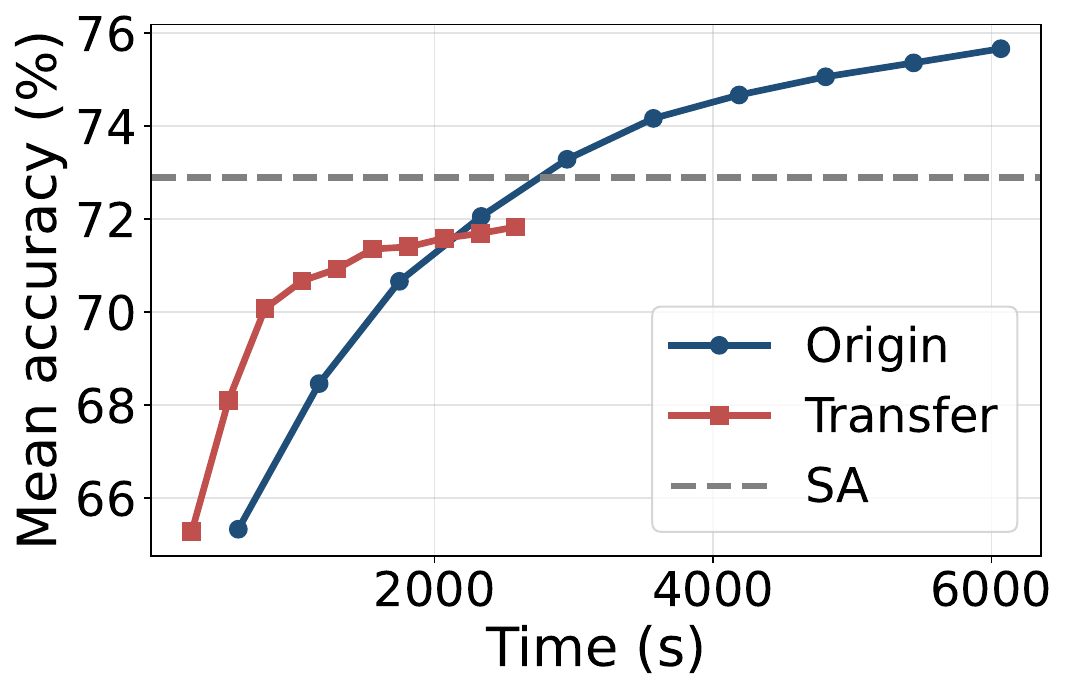}
        \caption{AdaMerging}
    \end{subfigure}
    \hfill
    \begin{subfigure}[b]{0.32\linewidth}
        \includegraphics[width=\linewidth]{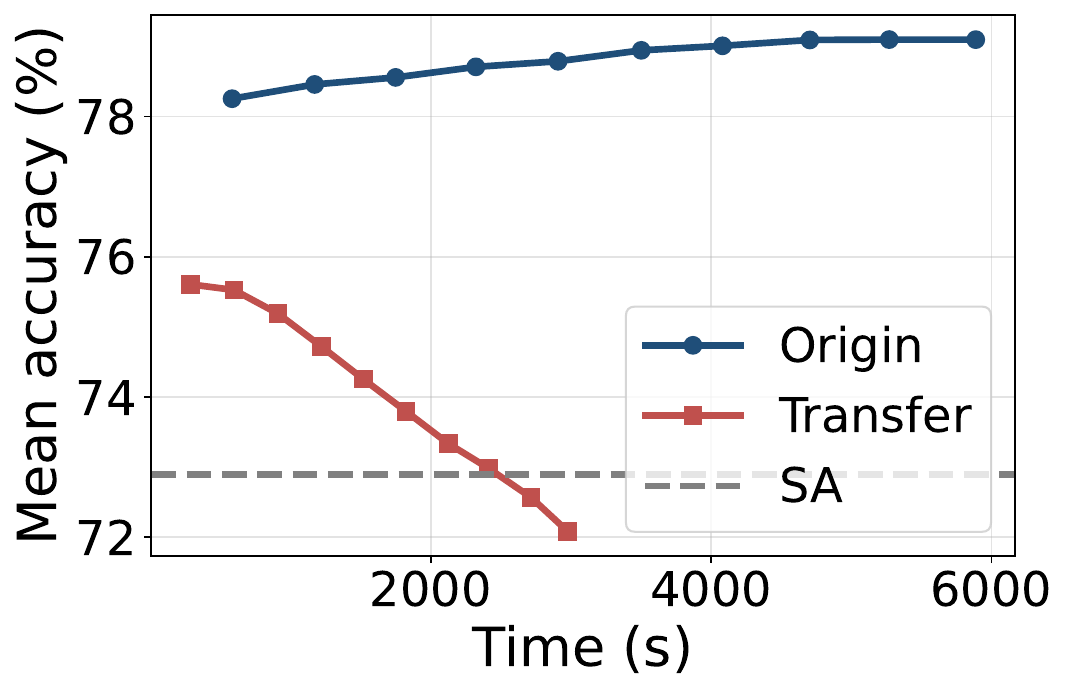}
        \caption{AdaMerging++}
    \end{subfigure}
    \hfill
    \begin{subfigure}[b]{0.32\linewidth}
        \includegraphics[width=\linewidth]{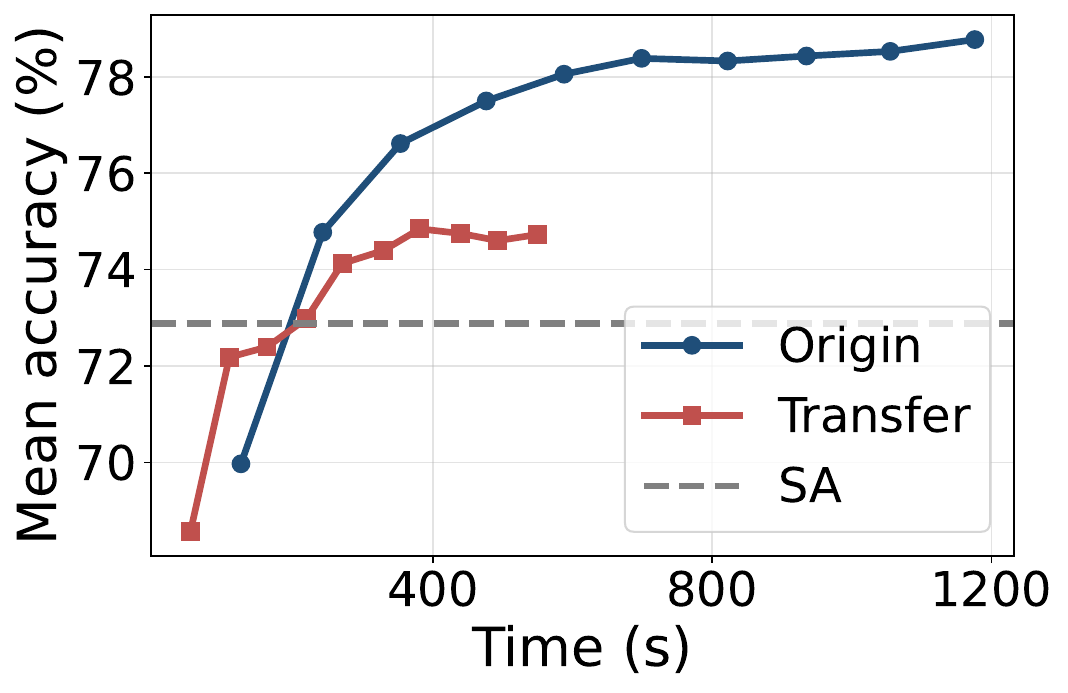}
        \caption{DivMerge}
    \end{subfigure}
    \caption{\textbf{Budget analysis on ViT-B/32 $\rightarrow$ ViT-L/14.} We compare direct search (Origin) against coefficient transfer ($\alpha$Transfer) under varying time budgets. Dashed lines indicate the simple average (SA) baseline. Top and bottom rows show evaluation- and gradient-based methods, respectively. Consistent with the SigLIP results, $\alpha$Transfer shifts the performance curves to the left, reaching comparable accuracy in a fraction of the time.}
    \label{fig:budget_b32_l14}
\end{figure}

\begin{figure}[h]
    \centering
    \begin{subfigure}[b]{0.32\linewidth}
        \includegraphics[width=\linewidth]{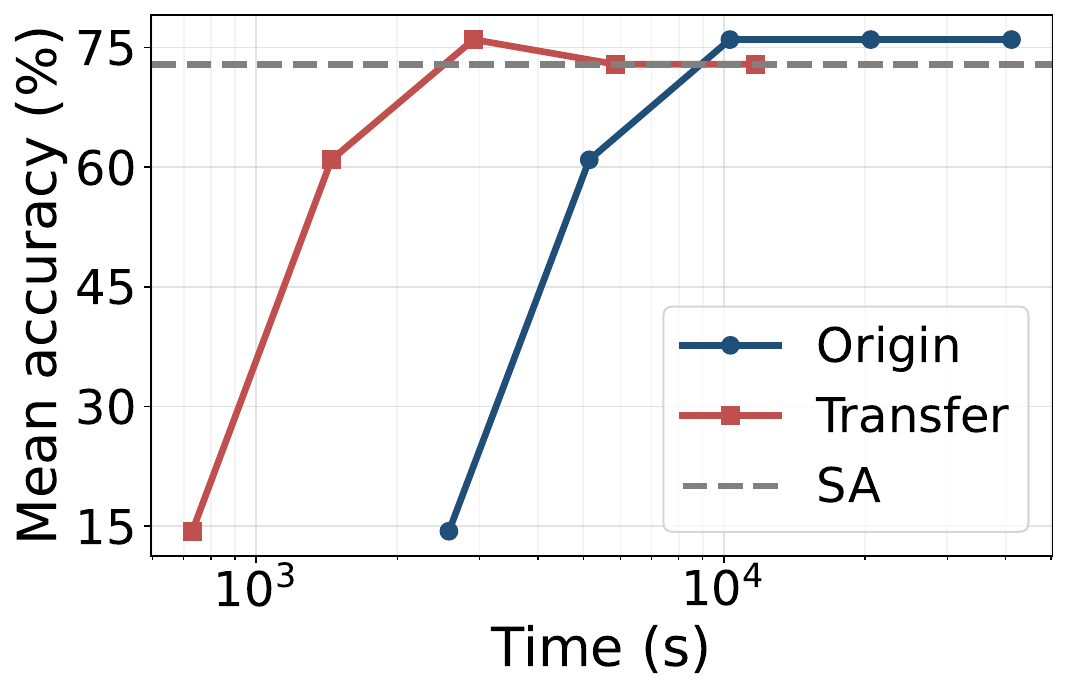}
        \caption{Task Arithmetic}
    \end{subfigure}
    \hfill
    \begin{subfigure}[b]{0.32\linewidth}
        \includegraphics[width=\linewidth]{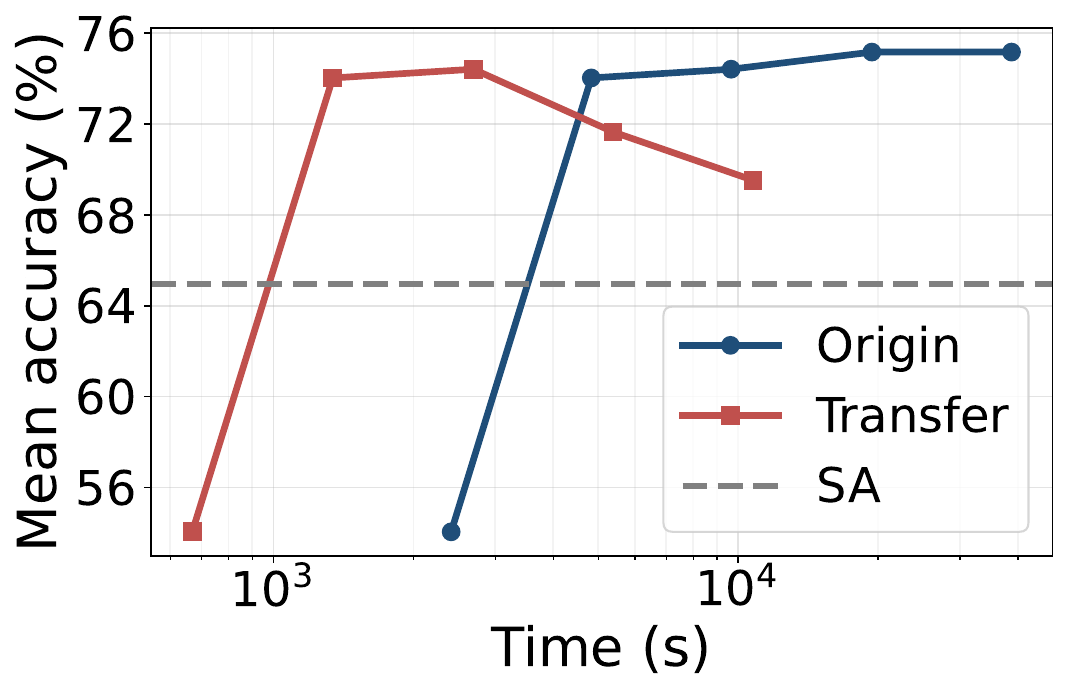}
        \caption{TIES}
    \end{subfigure}
    \hfill
    \begin{subfigure}[b]{0.32\linewidth}
        \includegraphics[width=\linewidth]{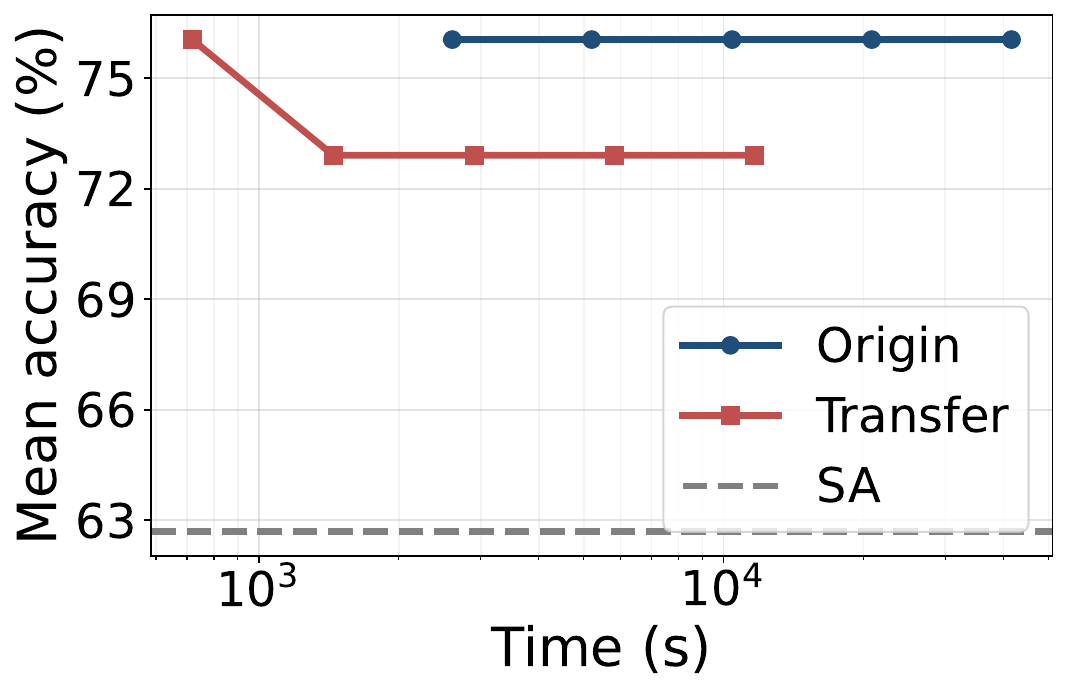}
        \caption{DARE}
    \end{subfigure}
    \vspace{0.5em}
    \begin{subfigure}[b]{0.32\linewidth}
        \includegraphics[width=\linewidth]{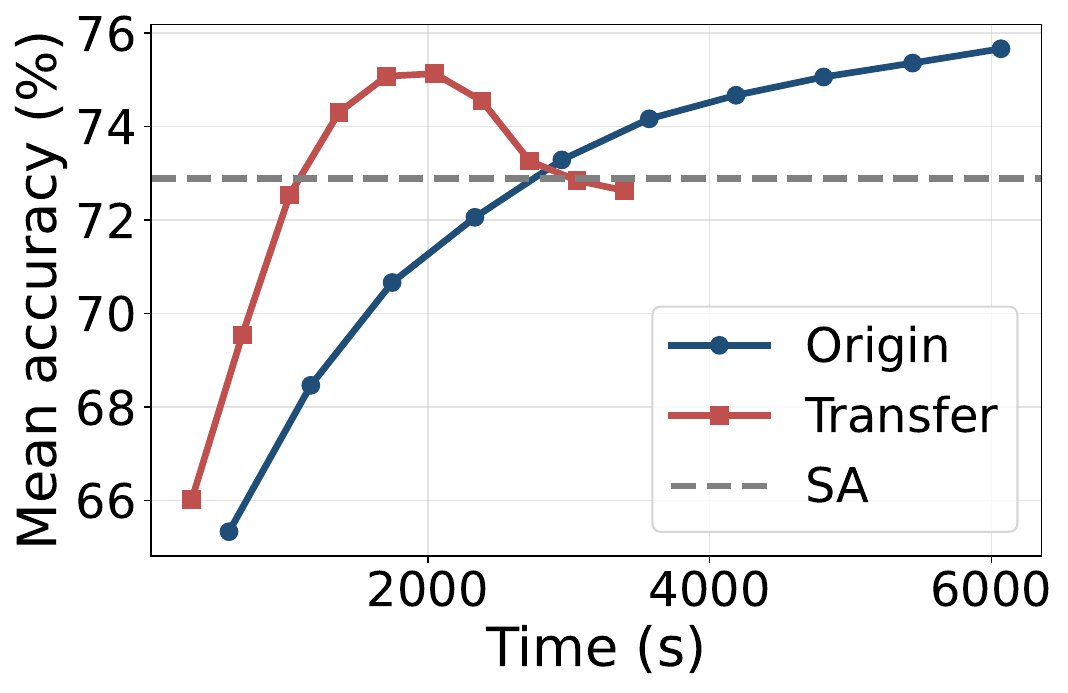}
        \caption{AdaMerging}
    \end{subfigure}
    \hfill
    \begin{subfigure}[b]{0.32\linewidth}
        \includegraphics[width=\linewidth]{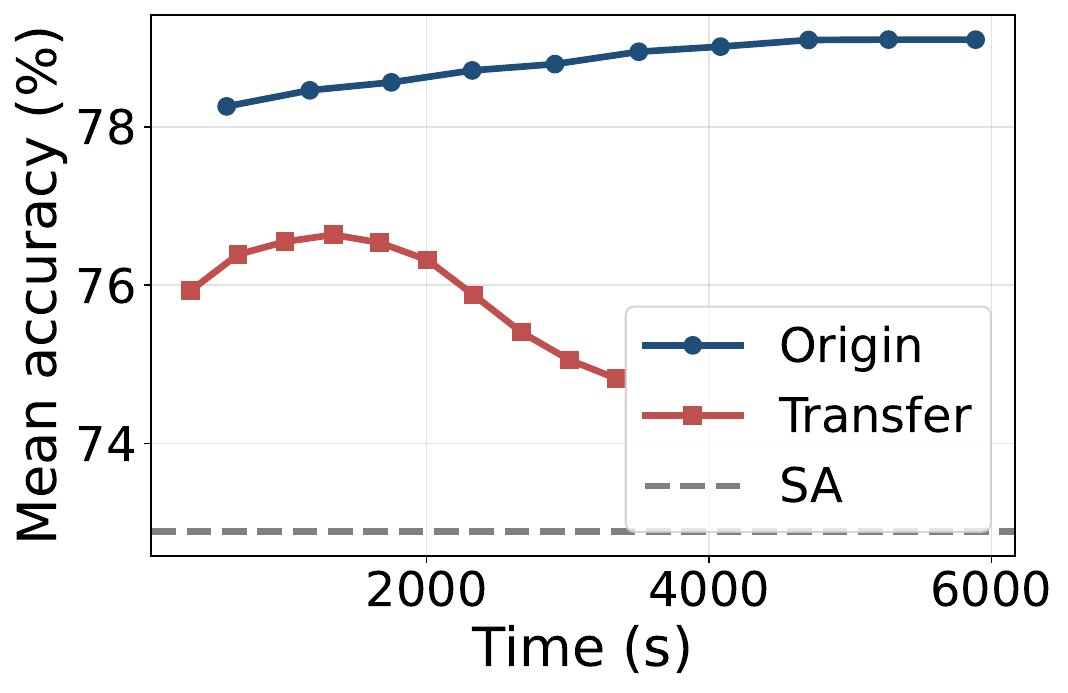}
        \caption{AdaMerging++}
    \end{subfigure}
    \hfill
    \begin{subfigure}[b]{0.32\linewidth}
        \includegraphics[width=\linewidth]{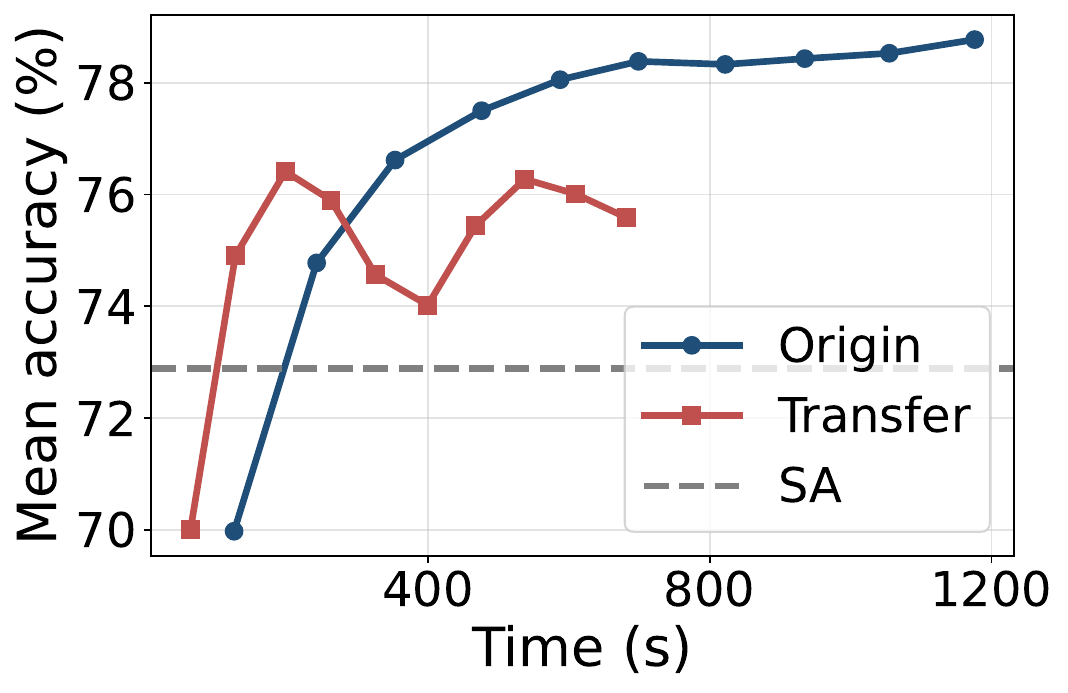}
        \caption{DivMerge}
    \end{subfigure}
    \caption{\textbf{Budget analysis on ViT-B/16 $\rightarrow$ ViT-L/14.} We compare direct search (Origin) against coefficient transfer ($\alpha$Transfer) under varying time budgets. Dashed lines indicate the simple average (SA) baseline. Top and bottom rows show evaluation- and gradient-based methods, respectively. Consistent with previous observations, $\alpha$Transfer shifts the performance curves to the left, reaching comparable accuracy in a fraction of the time.}
    \label{fig:budget_b16_l14}
\end{figure}

In this section, we provide the complete budget analysis curves for the two proxy$\to$target pairs not shown in the main paper: ViT-B/32 $\to$ ViT-L/14 (\cref{fig:budget_b32_l14}) and ViT-B/16 $\to$ ViT-L/14 (\cref{fig:budget_b16_l14}). The trends are consistent with those reported for SigLIP-Base $\to$ SigLIP-Large in the main paper: across all six merging methods, transferring search/training results from a smaller proxy model achieves accuracy comparable to (or surpassing) the simple-average baseline at a substantially lower computational budget.

\subsection{Effect of \textit{k} in AdaMerging++}
\label{app:failure_cases}

\begin{figure}[h]
    \centering
    \includegraphics[width=\linewidth]{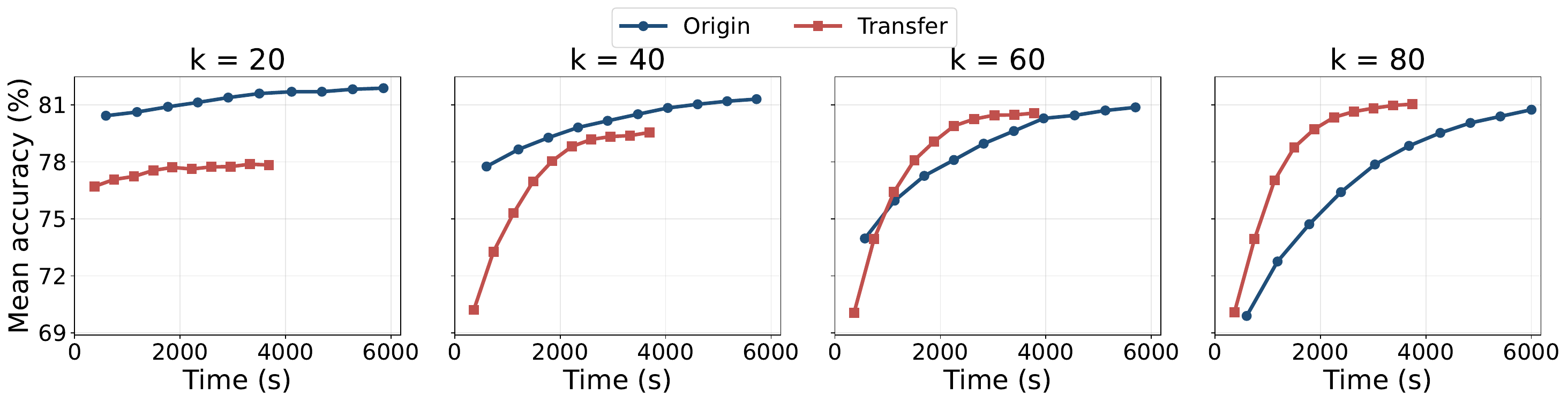}
    \caption{\textbf{Impact of sparsity on $\alpha$Transfer in AdaMerging++.} We compare exhaustive target optimization (Origin) against $\alpha$Transfer across varying TIES retention ratios $k \in \{20, 40, 60, 80\}$ on the SigLIP-Base $\rightarrow$ SigLIP-Large scaling chain. The x-axis represents the wall-clock training time. The effectiveness of $\alpha$Transfer is highly sensitive to the sparsity level: at extreme sparsity ($k=20$), direct target search achieves a higher performance ceiling. However, as the retention ratio increases ($k \ge 60$), $\alpha$Transfer dominates the compute-performance trade-off, achieving massive speedups while converging to a higher final accuracy than direct target search.}
    \label{fig:adamergepp_failure}
\end{figure}

Since AdaMerging++ incorporates TIES-style masking, its optimization dynamics are intrinsically tied to the sparsity parameter $k$ (the percentage of weights retained). To investigate the performance gap observed in our primary evaluation, we conduct an ablation study varying $k \in \{20, 40, 60, 80\}$ for the SigLIP Base-to-Large transfer, comparing direct target search (\emph{Origin}) against \textbf{$\alpha$Transfer} (Transfer) (\cref{fig:adamergepp_failure}).

The results indicate that the effectiveness of $\alpha$Transfer in AdaMerging++ is influenced by the sparsity level. Under higher sparsity settings ($k=20, 40$), heavy masking can enhance structural differences between proxy and target models, which moderately limits transferability and allows direct target search to achieve a higher performance ceiling. 

Conversely, under lower sparsity ($k=60, 80$), $\alpha$Transfer yields strong results. It demonstrates faster initial convergence and achieves comparable or higher final accuracy than direct target optimization. This observation suggests that $\alpha$Transfer remains a viable and efficient strategy for AdaMerging++ when a sufficient proportion of weights is retained during the optimization process.
 
\clearpage
\section{Per-task accuracy breakdown}

\subsection{Vision models}
\cref{tab:pertask_full} reports the per-task accuracy for each proxy $\to$ target pair, complementing the average scores reported in the main paper.
% Per-task accuracy tables (Appendix)
% Requires: booktabs, multirow, siunitx (or just use plain tabular)
% Compile with: pdflatex or any standard LaTeX engine

\newcommand{\ortr}[1]{\textit{#1}}  % style for Or/Tr label

% ============================================================
\begin{table}[p]
\centering
\caption{Per-task accuracy (\%) for each proxy$\to$target pair.
  \textbf{Or} = origin (target's best coefficients evaluated on target);
  \textbf{Tr} = transfer (proxy-best coefficients evaluated on target).}
\label{tab:pertask_full}
\setlength{\tabcolsep}{4pt}
\small
\newcolumntype{N}{>{\centering\arraybackslash}p{3.0em}}  % fixed-width centred numeric col
\begin{tabular}{ll NNNNNNNNN}
\toprule
\multicolumn{2}{l}{Method}
  & DTD & EuroSAT & FER & Food & GTSRB & RESISC & Cars & SUN & Avg \\
\midrule
\multicolumn{10}{l}{\textit{ViT-B/32 $\to$ ViT-L/14}} \\
\midrule
\multirow{2}{*}{Task Arithmetic}
  & Or & 59.63 & 89.07 & 51.67 & 91.50 & 77.53 & 85.40 & 79.59 & 73.48 & 75.98 \\
  & Tr & 60.53 & 87.11 & 50.35 & 92.62 & 73.34 & 85.59 & 81.10 & 73.77 & 75.55 \\
\multirow{2}{*}{TIES}
  & Or & 54.41 & 85.04 & 55.92 & 93.02 & 74.35 & 86.25 & 77.29 & 75.06 & 75.17 \\
  & Tr & 57.23 & 75.85 & 49.96 & 94.13 & 58.12 & 84.03 & 80.20 & 73.70 & 71.65 \\
\multirow{2}{*}{DARE}
  & Or & 60.05 & 88.74 & 51.11 & 92.04 & 76.67 & 85.65 & 80.39 & 73.72 & 76.05 \\
  & Tr & 60.59 & 87.07 & 50.32 & 92.63 & 73.33 & 85.56 & 81.18 & 73.77 & 75.56 \\
\multirow{2}{*}{AdaMerging}
  & Or & 59.68 & 89.93 & 49.39 & 85.90 & 91.83 & 81.24 & 75.96 & 71.41 & 75.67 \\
  & Tr & 58.14 & 97.67 & 47.41 & 87.80 & 63.49 & 72.79 & 76.05 & 71.30 & 71.83 \\
\multirow{2}{*}{AdaMerging++}
  & Or & 62.98 & 88.04 & 56.09 & 89.87 & 92.94 & 86.03 & 82.48 & 74.38 & 79.10 \\
  & Tr & 56.54 & 97.44 & 49.68 & 85.52 & 64.09 & 77.03 & 76.36 & 69.97 & 72.08 \\
\multirow{2}{*}{DivMerge}
  & Or & 72.55 & 89.70 & 60.64 & 86.75 & 89.90 & 78.43 & 82.34 & 69.86 & 78.77 \\
  & Tr & 69.10 & 97.33 & 40.50 & 89.59 & 68.54 & 81.89 & 79.39 & 71.50 & 74.73 \\
\midrule
\multicolumn{10}{l}{\textit{ViT-B/16 $\to$ ViT-L/14}} \\
\midrule
\multirow{2}{*}{Task Arithmetic}
  & Or & 59.63 & 89.07 & 51.67 & 91.50 & 77.53 & 85.40 & 79.59 & 73.48 & 75.98 \\
  & Tr & 60.53 & 81.93 & 47.21 & 93.31 & 62.05 & 83.78 & 81.36 & 72.94 & 72.89 \\
\multirow{2}{*}{TIES}
  & Or & 54.41 & 85.04 & 55.92 & 93.02 & 74.35 & 86.25 & 77.29 & 75.06 & 75.17 \\
  & Tr & 56.76 & 70.07 & 47.37 & 94.11 & 53.25 & 81.63 & 80.14 & 72.88 & 69.53 \\
\multirow{2}{*}{DARE}
  & Or & 60.05 & 88.74 & 51.11 & 92.04 & 76.67 & 85.65 & 80.39 & 73.72 & 76.05 \\
  & Tr & 60.69 & 81.81 & 47.19 & 93.31 & 62.10 & 83.83 & 81.40 & 72.95 & 72.91 \\
\multirow{2}{*}{AdaMerging}
  & Or & 59.68 & 89.93 & 49.39 & 85.90 & 91.83 & 81.24 & 75.96 & 71.41 & 75.67 \\
  & Tr & 60.21 & 66.04 & 45.37 & 90.51 & 97.89 & 70.89 & 74.06 & 76.07 & 72.63 \\
\multirow{2}{*}{AdaMerging++}
  & Or & 62.98 & 88.04 & 56.09 & 89.87 & 92.94 & 86.03 & 82.48 & 74.38 & 79.10 \\
  & Tr & 62.07 & 71.89 & 50.49 & 89.51 & 97.88 & 75.14 & 75.64 & 75.92 & 74.82 \\
\multirow{2}{*}{DivMerge}
  & Or & 72.55 & 89.70 & 60.64 & 86.75 & 89.90 & 78.43 & 82.34 & 69.86 & 78.77 \\
  & Tr & 78.03 & 68.33 & 49.29 & 89.52 & 96.44 & 73.92 & 75.29 & 73.84 & 75.58 \\
\midrule
\multicolumn{10}{l}{\textit{SigLIP-Base $\to$ SigLIP-Large}} \\
\midrule
\multirow{2}{*}{Task Arithmetic}
  & Or & 72.82 & 96.37 & 57.26 & 90.19 & 89.27 & 68.05 & 93.97 & 73.87 & 80.22 \\
  & Tr & 71.38 & 91.15 & 51.98 & 93.22 & 82.15 & 72.25 & 94.13 & 74.36 & 78.83 \\
\multirow{2}{*}{TIES}
  & Or & 66.91 & 96.67 & 61.94 & 90.25 & 82.95 & 58.05 & 90.47 & 72.73 & 77.50 \\
  & Tr & 67.39 & 95.59 & 59.15 & 91.53 & 81.37 & 59.87 & 91.17 & 73.05 & 77.39 \\
\multirow{2}{*}{DARE}
  & Or & 72.66 & 94.48 & 53.61 & 92.61 & 85.08 & 71.75 & 94.07 & 74.43 & 79.84 \\
  & Tr & 72.39 & 93.33 & 53.05 & 92.83 & 84.24 & 71.89 & 94.07 & 74.39 & 79.52 \\
\multirow{2}{*}{AdaMerging}
  & Or & 75.74 & 93.89 & 52.84 & 88.45 & 93.48 & 77.70 & 93.98 & 75.75 & 81.48 \\
  & Tr & 81.91 & 95.19 & 45.75 & 88.11 & 93.77 & 71.94 & 93.56 & 71.95 & 80.27 \\
\multirow{2}{*}{AdaMerging++}
  & Or & 77.02 & 95.96 & 50.33 & 90.96 & 94.01 & 76.02 & 94.48 & 76.23 & 81.88 \\
  & Tr & 74.36 & 96.93 & 49.53 & 80.28 & 96.79 & 63.89 & 92.33 & 68.57 & 77.83 \\
\multirow{2}{*}{DivMerge}
  & Or & 76.49 & 90.22 & 58.85 & 88.88 & 90.77 & 87.17 & 94.12 & 77.35 & 82.98 \\
  & Tr & 81.76 & 94.78 & 40.25 & 89.94 & 95.20 & 74.16 & 92.34 & 71.63 & 80.01 \\
\bottomrule
\end{tabular}
\end{table}

\subsection{Large language models}

\cref{tab:pertask_llm} reports the per-task accuracy for each LLM proxy $\to$ target pair, complementing the average scores reported in the main paper.

\begin{table}[p]
\centering
\caption{Per-task accuracy (\%) for each LLM proxy$\to$target pair.
  \textbf{Origin} = target's best coefficients evaluated on target;
  \textbf{Transfer} = proxy-best coefficients evaluated on target.}
\label{tab:pertask_llm}
\setlength{\tabcolsep}{5pt}
\small
\newcolumntype{N}{>{\centering\arraybackslash}p{4.0em}}
\begin{tabular}{llNNNNN}
\toprule
\multicolumn{2}{l}{Method}
  & DDXPlus & Banking77 & Usefulness Judge & IFEval & Avg \\
\midrule
\multicolumn{7}{l}{\textit{Qwen3-0.6B $\to$ Qwen3-1.7B}} \\
\midrule
\multirow{2}{*}{Task Arithmetic}
  & Origin
    & 85.71 & 95.70 & 79.20 & 45.13 & 76.44 \\
  & Transfer
    & 80.78 & 93.50 & 77.60 & 46.64 & 74.63 \\
\multirow{2}{*}{TIES}
  & Origin
    & 8.56 & 77.10 & 82.40 & 65.34 & 58.35 \\
  & Transfer
    & 7.94 & 79.60 & 79.60 & 63.28 & 57.61 \\
\multirow{2}{*}{TIES-Pairwise}
  & Origin
    & 75.62 & 92.30 & 72.00 & 44.67 & 71.15 \\
  & Transfer
    & 72.17 & 90.40 & 72.80 & 46.09 & 70.37 \\
\midrule
\multicolumn{7}{l}{\textit{Qwen3-0.6B $\to$ Qwen3-4B}} \\
\midrule
\multirow{2}{*}{Task Arithmetic}
  & Origin
    & 91.21 & 92.00 & 78.40 & 42.51 & 76.03 \\
  & Transfer
    & 91.21 & 92.00 & 78.40 & 42.51 & 76.03 \\
\multirow{2}{*}{TIES}
  & Origin
    & 36.17 & 93.00 & 80.00 & 54.99 & 66.04 \\
  & Transfer
    & 35.77 & 93.00 & 81.20 & 51.05 & 65.26 \\
\multirow{2}{*}{TIES-Pairwise}
  & Origin
    & 89.74 & 90.10 & 76.00 & 38.94 & 73.69 \\
  & Transfer
    & 89.74 & 90.10 & 76.00 & 38.94 & 73.69 \\
\bottomrule
\end{tabular}
\end{table}

\end{document}